%% file: main.tex
\documentclass[10pt,twocolumn,letterpaper]{article}

\usepackage{preprint}             

\input{preamble}

\definecolor{blue}{rgb}{0.21,0.49,0.74}
\usepackage[pagebackref,breaklinks,colorlinks,allcolors=blue]{hyperref}

\title{\methodname: Full-Body Unified Motion Prior for Body and Hands via Diffusion}

\author{\fontsize{12pt}{\baselineskip}\selectfont
        Enes Duran\textsuperscript{1, 2} \quad Nikos Athanasiou\textsuperscript{1} \quad Muhammed Kocabas\textsuperscript{3} \quad  Michael J. Black\textsuperscript{1} \\ 
        \fontsize{12pt}{\baselineskip}\selectfont
        Omid Taheri\textsuperscript{1} \\
        \fontsize{11pt}{\baselineskip}\selectfont
        $^1$ Max Planck Institute for Intelligent Systems, ~
        $^2$ University of T\"ubingen ~ 
        $^3$ Meshcapade GmbH, Germany
    }
    
\begin{document}

\input{sections/teaser}    
\maketitle
\input{sections/0_abstract}
\input{sections/1_introduction}
\input{sections/2_related_work}

\input{sections/3_method}

\input{sections/4_applications}

\input{sections/5_experiments}

\input{sections/6_conclusion}
{\small
 \bibliographystyle{ieeenat_fullname.bst}
 \bibliography{main.bib}}

\input{sections/supmat}

\end{document}

%% file: preamble.tex
\usepackage[dvipsnames]{xcolor}

\usepackage{array}
\usepackage{xcolor} 
\usepackage{amsmath}
\usepackage{graphicx}
\usepackage{makecell}
\usepackage{mathtools}
\usepackage{natbib}

\newcommand{\methodname}{\textit{FUSION}\xspace}
\newcommand{\first}[1]{\textbf{#1}}

\usepackage[acronym]{glossaries}

\newacronym{smpl-x}{SMPL-X}{SMPL eXpressive}
\newacronym{gt}{GT}{Ground Truth}
\newacronym{ar}{AR}{Augmented Reality}
\newacronym{vr}{VR}{Virtual Reality}
\newacronym{llm}{LLM}{Large Language Models}
\newacronym{hoi}{HOI}{Human--Object Interaction}
\newacronym{sota}{SOTA}{State-Of-The-Art}
\newacronym{lbs}{LBS}{Linear Blend Skinning}

%% file: sections/teaser.tex
\twocolumn[{%
	\renewcommand\twocolumn[1][]{#1}%
	\maketitle   
	\begin{center}
    \vspace{-0.4cm} 
		\newcommand{\teaserwidth}{\textwidth}

		\centerline{
			\includegraphics[width=\teaserwidth,clip]{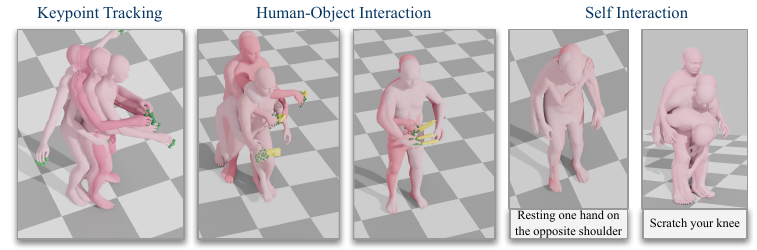}
		}
        \vspace{-0.4cm} 
		\captionof{figure}{Our method, \methodname, enables full-body motion synthesis, including detailed hand motion. Using our optimization framework, we achieve realistic motions for Keypoint Tracking (left), Human--Object Interaction (middle), and Self-Interaction (right) tasks. The initial frames have lighter colors and they get darker over time. To see the motions please watch the supplementary video.}  
		\label{fig:teaser}
        \vspace{-0.3cm} 
	\end{center}%
}]

%% file: sections/0_abstract.tex
\begin{abstract}
\vspace{-0.3cm}

Hands are central to interacting with our surroundings and conveying gestures, making their inclusion essential for full-body motion synthesis. Despite this, existing human motion synthesis methods fall short: some ignore hand motions entirely, while others generate full-body motions only for narrowly scoped tasks under highly constrained settings. A key obstacle is the lack of large-scale datasets that jointly capture diverse full-body motion with detailed hand articulation. While some datasets capture both, they are limited in scale and diversity. Conversely, large-scale datasets typically focus either on body motion without hands or on hand motions without the body. To overcome this, we curate and unify existing hand motion datasets with large-scale body motion data to generate full-body sequences that capture both hand and body. We then propose the first diffusion-based unconditional full-body motion prior, \methodname, which jointly models body and hand motion. Despite using a pose-based motion representation, \methodname surpasses \gls{sota} skeletal control models on the Keypoint Tracking task in the HumanML3D dataset and achieves superior motion naturalness. Beyond standard benchmarks, we demonstrate that \methodname can go beyond typical uses of motion priors through two applications: (1) generating detailed full-body motion including fingers during interaction given the motion of an object, and (2) generating Self-Interaction motions using an \gls{llm} to transform natural language cues into actionable motion constraints. For these applications, we develop an optimization pipeline that refines the latent space of our diffusion model to generate task-specific motions. Experiments on these tasks highlight precise control over hand motion while maintaining plausible full-body coordination. The code will be public.
\vspace{-0.6cm}
\end{abstract}

%% file: sections/1_introduction.tex
\section{Introduction}
\label{sec:intro}

Generating realistic human motion is essential for interactive applications spanning Computer Graphics,Computer Vision, Robotics, and \gls{ar}/\gls{vr}. A key aspect of realism in these settings is interaction, whether with objects, other people, one’s own body, or the user in a virtual environment. Even a simple act like \textit{drinking water from a glass} combines \gls{hoi} (grasping and manipulating the glass) with Self-Interaction (bringing it to the mouth). This requires coordinated full-body movement: torso leaning, arm and elbow adjustment to avoid collisions, anatomically correct grasps, subtle wrist rotations, and head motion. Any unnatural artifact (\eg finger penetration) immediately breaks immersion.

Because such tasks demand different body parts \textit{working in tandem}, it is crucial to capture both body and hand movements within a unified framework. Generating such motions directly from task descriptions or constraints is highly challenging due to the high-dimensional pose space, complex anatomical dependencies, the need for temporal consistency, and the precise coordination required for interactions. To address these challenges, many approaches rely on motion priors, \ie learned models that provide a distribution over plausible human motions, simplifying the task of generating realistic sequences. Yet, most existing models focus solely on body motion, neglecting hand articulation~\cite{rempe2021humor, athanasiou2024motionfix, li2024unimotion, he2022nemf, wandr, TEACH:3DV:2022, SINC:2023, TLControl, MDM, omomo}. This leads to a disconnect between body and hand behavior, limiting the coherence of motions. While recent methods attempt to address this gap, they remain narrowly  scoped~\cite{ron2025hoidinihumanobjectinteractiondiffusion, ghosh2022imos, taheri2024grip}, targeting specific tasks such as \gls{hoi} with relatively static body motions and under strong constraints \eg predefined text descriptions, initial body pose, or body motion without hands. 

Despite its importance, modeling full-body motions remains largely understudied. A significant barrier is the lack of large-scale datasets that simultaneously capture hand, head, and body motions. While there exist datasets that capture both hand and body, these typically contain very limited body motion~\cite{GRAB, ARCTIC, TCDHands, emage1},
making them insufficient for training general-purpose motion priors. In contrast, large-scale datasets capturing rich body motions typically lack detailed hands~\cite{AMASS, omomo}, while others focus narrowly on hand articulation without body motion~\cite{Reinterhand, InterHand2.6M, HOT3D}. As a result, no single dataset offers the scale and diversity of natural full-body motion with hands. 

Our key insight is to merge this limited full-body data with large-scale body-only datasets and hand-only datasets. We combine detailed local hand motions with broader body motions to obtain additional full-body sequences. While these augmented sequences are not correlated at the sequence level, they still serve effectively as training data for a motion prior, which models plausible human motions. During downstream motion generation, observation constraints (\eg trajectories, contacts, or task-specific guidance) ensure consistency with the target scenario. We empirically validate this strategy through ablation studies, showing that the joint prior outperforms separate body-only and hand-only priors.



Building on this idea, we propose \methodname, a novel unconditional denoising diffusion-based full-body motion prior that jointly models body and hand dynamics. Training an unconditional model is advantageous as it allows leveraging a broad range of motion data without being limited by specific conditioning inputs such as text descriptions, audio signals, or object-related information. By removing these constraints during training, the model learns a more general and natural representation of human motion. The model can later be adapted or fine-tuned for various targeted applications. To enable such adaptation, we build upon the diffusion noise optimization method DNO~\cite{DNO}, extending it to handle flexible constraints from diverse sources, such as keypoint trajectories, high-level instructions from external planners such as \gls{llm}s, and object grasping algorithms. This extension broadens the applicability of DNO to unified full-body motion control, capturing both coarse body dynamics and fine-grained finger articulations.



We demonstrate the effectiveness and flexibility of \methodname through extensive quantitative and qualitative experiments on Keypoint Tracking, \gls{hoi}, and Self-Interaction tasks (see~\cref{fig:teaser}, \cref{tab:humanml3d_all}, \cref{fig:qualitative_results}). In particular, we show that \methodname can accurately track spatial trajectories across various body parts. Additionally, given the motion of an object as a constraint, \methodname can be used to generate detailed full-body motions that naturally conform to the object's movement. Lastly, we provide natural language commands (e.g., \textit{touch your head}) to an \gls{llm}, which extracts contact-related cues in the form of body vertex constraints over time. \methodname then uses these constraints to synthesize Self-Interaction motions that follow given commands.

Overall, our work advances full-body motion synthesis by providing a unified framework that seamlessly integrates coarse motion generation with fine-grained interactive control. In summary, our contributions are as follows: 

\begin{itemize} 
\item We propose a novel way to combine and augment heterogeneous data sources for hands and body, and use them to train \methodname, the first unconditional diffusion-based motion prior that synthesizes full-body motions in a unified framework.
\item We extend a baseline diffusion noise optimization (DNO) framework with task-specific modifications, enabling our trained motion prior to adapt to diverse downstream applications such as \gls{hoi} and Self-Interaction.
\item We propose a novel pipeline that uses \gls{llm}s to convert natural language cues into constraints to generate realistic Self-Interaction motions. Through the optimization we generated a small Self-Interaction dataset that can be useful in other downstream tasks.
\end{itemize}
  

%% file: sections/2_related_work.tex
\section{Related Work}
\label{sec:related_work }

\subsection{Human Motion Synthesis}
\label{sec:human_motion_synthesis }

Motion synthesis methods can be categorized into unconditional and conditional approaches. Unconditional methods generate human motion without control signals. HuMoR~\cite{rempe2021humor} uses an autoregressive VAE architecture~\cite{kingma2013auto}, but this design complicates test-time optimization. NeMF, alleviates this problem through sequential modeling of the human motion~\cite{he2022nemf}. NeMF separates global from local motion with a lightweight representation, performing well in motion inbetweening and interpolation but is limited in diverse motion synthesis. HMP~\cite{HMP} extends NeMF to hand motions, but it cannot capture coordinated body and hand dynamics, underscoring the need for a unified motion prior. 

In contrast, conditional approaches focus on generating motions using specific control inputs. These methods use various conditioning signals such as text descriptions~\cite{PriorMDM, zhang2023generating, tevet2022motionclip, li2024unimotion, athanasiou2024motionfix, petrovich23tmr, MDM, SINC:2023, ron2025hoidinihumanobjectinteractiondiffusion}, audio~\cite{AI_Choreographer, tseng2022edge}, trajectories~\cite{rempeluo2023tracepace, GMD}, action labels~\cite{ACTOR:ICCV:2021}, and object-related information~\cite{ghosh2022imos, taheri2024grip, xu2023interdiff, zhang22couch, Pi_2023_ICCV, omomo, christen2024diffh2o, taheri2021goal}. While these approaches enable context-aware motion synthesis, they each require specific data types, limiting the use of available datasets. By contrast, \methodname leverages heterogeneous data sources without being bound to specific conditioning signals, while still allowing for flexible control through optimization.


\subsection{Spatial Control for Human Motion Synthesis}
\label{sec:spatial_control}

Recent research has focused on controlling motion through spatial cues, particularly joint positions over time. GMD~\cite{GMD} uses a two-stage diffusion approach but controls only 2D pelvis trajectories. OmniControl~\cite{omnicontrol} offers flexible spatial control over various joints using spatial and realism guidance, though it struggles with accurate trajectory alignment. TLControl~\cite{TLControl} achieves better trajectory control through latent code optimization of motion VQ-VAE. However, these methods discard hands and operate in the joint space, requiring post-processing to convert motion into pose space. WANDR~\cite{wandr} addresses this by working directly in pose space and introducing \textit{intention features}, but has no hand articulation and is limited in controlling multiple joints. \methodname advances beyond these approaches by providing flexible control over both body and hand motions in a single framework, operating directly in pose space while avoiding the limitations of autoregressive models through our  architecture and noise optimization technique.

\subsection{Controlling Diffusion Models}
\label{sec:controlling_diffusion_models}

Diffusion models show strong performance in synthesis tasks but present challenges in control. Existing control methods for diffusion models include classifier guidance~\cite{Nichol2021ImprovedDD, song2021scorebased}, classifier-free guidance~\cite{ho2021classifierfree, Rombach2021HighResolutionIS, photorealistic_t2i, zhang2024rohm, karras2024guiding}, score distillation~\cite{poole2022dreamfusion}, and diffusion noise optimization~\cite{DNO, DOODL}. DNO~\cite{DNO} has successfully applied noise optimization to tasks such as motion inbetweening, dense optimization, keyframe interpolation, and obstacle avoidance. Building on this work, we control our full-body motion model through noise optimization across various tasks. Our approach extends these capabilities by integrating flexible constraints derived from diverse sources ranging from keypoint trajectories to high-level instructions from \gls{llm}s or object grasping algorithms. This enables precise control over both coarse body movements and fine-grained finger articulations within a unified framework.

%% file: sections/3_method.tex
\section{Method}

This section introduces \methodname, the first generic motion prior capable of synthesizing full-body motions, describes our novel data fusion strategy that combines heterogeneous hand and body motion datasets, and presents our flexible optimization framework that refines the latent space of our diffusion model using task-specific constraints.
 
\input{figures/tex_files/main_body/method_overview}

\subsection{Curating the Dataset}
\label{sec:dataset_curation}

A major limitation in full-body motion synthesis is the scarcity of datasets that include both detailed hand and body motion. Existing datasets that capture both typically focus on \gls{hoi} or gesture synthesis, where the body remains relatively stable with minimal movement~\cite{emage1, GRAB, ARCTIC}. This limitation has long been a key obstacle to developing a comprehensive motion prior capable of synthesizing realistic hand and body motions together. To overcome this challenge, we employ a method that effectively leverages available data for both hand and body motion.

To obtain body motions, we compile data from AMASS, ARCTIC, OMOMO, SAMP, and BEAT2~\cite{AMASS, ARCTIC, omomo, SAMP, emage1}. To augment this curated dataset, we apply flipping to both pose and translation, effectively doubling the number of samples while ensuring the motions remain plausible. This straightforward augmentation increases the diversity of body motion sequences without compromising their naturalness. After processing, we have \(145,100\) sequences, each with \(120\) timeframes.

For hand motions, we combine data from ARCTIC, TCDHands, GRAB, HOT3D, MOYO, InterHand2.6M, and Re:InterHand~\cite{ARCTIC, TCDHands, GRAB, HOT3D, MOYO, InterHand2.6M, Reinterhand}. Our augmentation process begins by flipping local left-hand motion sequences to generate right-hand counterparts, while right-hand motions are kept in their original form. This pose-flipping strategy enhances data diversity without compromising motion plausibility. Additionally, we apply time-reversal augmentation by reversing the temporal order of hand motions, which further doubles the dataset while preserving motion realism. From this resulting set of local right-hand motion sequences, we randomly sample segments and integrate them into the full-body motion. For left-hand motions, pose flipping is applied in a post-processing step to ensure consistency. When the body motion is from a full-body dataset that already includes hand motions (\eg, GRAB, ARCTIC, BEAT2, TCDHands), we instead retain the original hand segments to better preserve hand–body dynamics. 

In total, we have \(41,148\) sequences, each with \(120\) timeframes. Overall, the augmentation strategies improve the model's ability to generalize to diverse dynamic hand and body motions. Our test split consists of the test sets from the ARCTIC and GRAB datasets~\cite{ARCTIC, GRAB}. In total, our dataset comprises \(17.4\)M timeframes at \(30\) FPS, with each sequence containing \(120\) timeframes. This curated dataset provides a rich resource for full-body motion synthesis. \

 
A key concern when merging body and hand motions is the risk of self-penetration. To address this, we always retain the wrist orientation from the body pose and use only the local hand articulation from the hand motion when merging datasets. Keeping the wrist orientation consistent with the body pose reduces the risk of self-penetration. In addition, we check whether copying local hand motions introduces further self-penetrations and apply penetration-based filtering, discarding physically impossible motions. For dataset ratios and motion merging visualization, we refer readers to Sup.Mat..


\subsection{Motion Representation}
\methodname learns 3D human motion by extracting features from \gls{smpl-x}~\cite{SMPL-X:2019} motion parameters. Recent works on motion editing and synthesis have demonstrated that predicting canonicalized trajectories is advantageous for both synthesis and editing tasks~\cite{athanasiou2024motionfix, omomo, TLControl, zhang2024rohm}. Building on this observation, we predict the canonicalized trajectory values. A motion sequence \(\boldsymbol{X}\!=\!\{x^{1}, \dots, x^{N}\}\!\in\!\mathbb{R}^{N \times D}\) consists of a sequence of feature vectors \(x^{i} \in \mathbb{R}^{D}\): 
\[
x^{i} \coloneqq \{\tau^{i}, \phi^{i}, \gamma^{i}, \theta^{i}, J^{i}, f^{i}\}.
\]
Here, \(\tau^{i} \in \mathbb{R}^{3}\) is the canonicalized root velocity, \(\phi^{i} \in \mathbb{R}^{6}\) is the root orientation with respect to the XY plane, \(\gamma^{i} \in \mathbb{R}^{6}\) represents the root angular velocity with respect to the Z axis, \(\theta^{i} \in \mathbb{R}^{324}\) is the local SMPL-X pose in 6D representation~\cite{6D_representation}, \(J^{i} \in \mathbb{R}^{165}\) is the corresponding joint locations, and \(f^{i} \in [0,1]^{4}\) are the predicted contact probabilities of the foot joints. \(N=120\) is the sequence length, \(D=508\) is the feature dimension.

\subsection{\methodname Formulation}
Previous methods have shown the effectiveness of diffusion models for motion generation~\cite{MDM, omnicontrol, omomo, athanasiou2024motionfix, PriorMDM}. Motivated by these results, we train \methodname as an unconditional denoising diffusion model with fixed sequence lengths. The model learns the reverse diffusion process of gradually denoising \(X_{t}\), starting from pure Gaussian noise \(X_{T}\): \(P_{\theta_M} \left(\boldsymbol{X}_{t-1} \mid \boldsymbol{X}_t\right) = \mathcal{N}\left(\boldsymbol{\mu}_t(\theta_M), \left(1 - \alpha_t\right) \boldsymbol{I}\right),\) where \(\theta_M\) is the denoising diffusion model parameters, \(\boldsymbol{X}_t \in \mathbb{R}^{N \times D}\) represents the motion state at the \(t^\text{th}\) step of the noising process, and \(T=300\) denotes the total number of denoising steps. The hyperparameters \(\alpha_t \in (0,1)\) progressively decrease to \(0\) as \(t\rightarrow T \). Inspired by \cite{MDM}, our model directly estimates the final clean motion \(\hat{\boldsymbol{X}}_0(\theta_M) \coloneqq M(\boldsymbol{X}_t, t; \theta_M) \),
where \(M\) is the motion denoiser model. The model is trained through the reconstruction, geometry, and foot skating losses:
\begin{equation}
\begin{gathered}
\mathcal{L}_{\text{recon}} \coloneqq E_{t \sim \text{DiscreteUniform}[1, T]}  \|\boldsymbol{X}_0-\hat{\boldsymbol{X}}_0\|_2^2, \\
\mathcal{L}_{\text {geo}} \coloneqq \frac{1}{N} \sum_{i=1}^N\left\|FK(x_0^i)-FK(\hat{x}_0^i)\right\|_2^2, \\
\mathcal{L}_{\text {foot}} \coloneqq \frac{1}{N-1} \sum_{i=1}^{N-1}\left\|\left(FK(\hat{x}_0^{i+1})-FK(\hat{x}_0^i)\right)\cdot f^{i}\right\|_2^2,
\end{gathered}
\end{equation}
where \(\boldsymbol{X}_0\) denotes the \gls{gt} motion sequence, FK(\(\cdot\)) denotes the forward kinematics function of the \gls{smpl-x} body model~\cite{SMPL-X:2019}, \(x_{0}^{i}\) is the \gls{gt} pose sample, \(\hat{x}_{0}^{i}\) is the denoised pose sample, and \(f^{i} \in \{0, 1\}\) is the binary contact prediction at frame \(i\). The term \(\mathcal{L}_{\text{geo}}\) ensures that the predicted joints are close to those computed via forward kinematics, and \(\mathcal{L}_{\text{foot}}\) mitigates foot skating.

\subsection{\methodname Implementation details}

We base our model on MotionFix~\cite{athanasiou2024motionfix} and modify its transformer encoder architecture . The architecture consists of \(19.7\)M trainable parameters. The learning rate is set to \(10^{-4}\), and the batch size is \(32\), with a total of \(300{,}000\) training steps. We use the AdamW~\cite{Loshchilov2017DecoupledWD} optimizer. The forward diffusion step count \(T\) is set to \(300\). The loss hyperparameters are \(\lambda_{\text{recon}} = 1\), \(\lambda_{\text{geo}} = 1\), and \(\lambda_{\text{foot}} = 1\). For more implementation details, please refer to the project webpage.

\subsection{Diffusion Noise Optimization}

Unconditional motion priors allow the model to learn a diverse latent space without requiring labeled or task-specific training data. However, without conditioning or constraints, they are not directly useful for specific downstream tasks such as \gls{hoi}, Self-Interaction, or other forms of motion that depend on external conditions (see \cref{sec:application}). To make them practical, it is essential to control the generated motion so that it aligns with task-specific requirements. This calls for a framework to control the motion. Recently, DNO~\cite{DNO} has demonstrated that optimizing diffusion noise is an effective way to perform motion synthesis for various tasks.


DNO~\cite{DNO} treats the diffusion noise \(x_T \sim \mathcal{N}(0, I)\) as the latent variable and generates motion with the full sampling process of the trained diffusion model \(M\), resolved using the ODE formulation of DDIM~\cite{song2021denoising}. The optimization is then defined over \(x_T\), with gradients propagating through all denoising steps:
\[
x_T^* \coloneqq \arg \min_{x_T} \left\{ \mathcal{L}_{\text{opt}}(\mathrm{ODE}(M, x_T)) + \mathcal{L}_{\mathrm{decorr}}(x_T) \right\}
\]
where the final output will be \(x^* = \mathrm{ODE}(M, x_T^*)\). Here,
\(\mathcal{L}_{\mathrm{decorr}}\) is a decorrelation regularizer~\cite{StyleGAN}
that encourages \(x_T\) to remain within the Gaussian prior. \(\mathcal{L}_{\mathrm{opt}}\) includes several loss terms to guide the motion towards both realism and task compliance:
\begin{equation}
\begin{gathered}
\mathcal{L}_{\text{opt}} \coloneqq \lambda_{\text{lk}} \mathcal{L}_{\text{lk}} + \lambda_{\text{foot}} \mathcal{L}_{\text{foot}} + \lambda_{\text{ch}} \mathcal{L}_{\text{ch}} + \lambda_{\text{close}} \mathcal{L}_{\text{close}},
\end{gathered}
\end{equation}
where \(\mathcal{L}_{\text{lk}}\) encourages the diffusion noise to remain close to zero (the prior mean), \(\mathcal{L}_{\text{ch}}\) penalizes foot deviation from ground when in contact, and \(\mathcal{L}_{\text{close}}\) enforces tracking of specified keypoints over time. Specifically, let \( c^{k}_{j} \in \mathbb{R}^3 \) denote the target location for the joint (or vertex) \(j\) at the keyframe \(k\), and let \(\hat{c}^{k}_{j}\) be its computed location from the denoised motion \( \hat{\boldsymbol{x}}_{0}\). We then define:
\begin{equation}
\begin{gathered}
\mathcal{L}_{\text{lk}} \coloneqq \frac{1}{N} \sum_{i=1}^N\left\| x^{i}_{T} \right\|_2^2, \\
\mathcal{L}_{\text{ch}} \coloneqq \frac{1}{N-1} \sum_{i=1}^{N-1}\left\|FK(\hat{x}_0^i)_{z} \cdot f^{i}\right\|_1, \\
\mathcal{L}_{\text {close}}(\mathbf{x}, O) \coloneqq \frac{1}{|O|} \sum_{(j, k) \in O}\left\|\hat{\mathbf{c}}_j^k(\hat{\boldsymbol{x}}_{0})-\mathbf{c}_j^{k}\right\|_1,
\end{gathered}
\end{equation}
Here, \(O\) denotes the observed set containing \((j, k)\) pairs corresponding to target joints and keyframes determined by the task. \(\mathcal{L}_{\text{opt}} \) can include other types of losses, such as penetration loss or contact loss, depending on the specific application. \cref{fig:method_overview} shows an overview of our optimization setup and different constraints from downstream tasks. For hyperparameter values for the optimization, we refer to Sup.Mat.. 


%% file: figures/tex_files/main_body/method_overview.tex
\begin{figure}
\centerline{\includegraphics[width=0.98\linewidth]{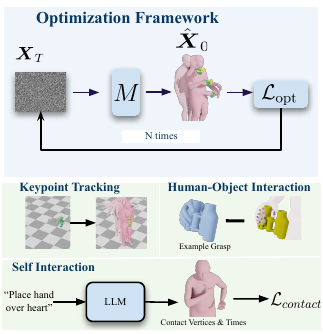}}
    \vspace{-0.3cm}
    \caption{\textbf{Method Overview:} Our framework (top) refines diffusion noise through an optimization process that incorporates multiple control signals (middle, bottom). These include example grasps (middle-right), where the difference between human hand joints (pink dots) and grasping hand joints (green dots) guides the optimization. Additionally, joint-based control signals (middle-left picture, green points) serve as loss functions to enforce precise motion constraints. Finally, high-level plans from an LLM (bottom) enable motion synthesis that adheres to specified contact vertices and timesteps, ensuring task-driven and contextually appropriate interactions. For each application our framework leverages backpropagation through the motion denoiser \(M\) to iteratively update the noise. After optimization, the refined noise \(X_{T}\) is decoded to produce the final motion sequence.}
    \vspace{-0.5cm}
    \label{fig:method_overview}
\end{figure}

%% file: sections/4_applications.tex
\section{Applications}
\label{sec:application}

\input{figures/tex_files/main_body/qualitative_results}

\methodname unlocks a broad range of applications by providing a unified motion prior that synthesizes full-body and finger motions. By combining body and finger articulation, our method captures fine-grained interactions, and using our diffusion noise optimization framework, it can be efficiently guided when provided with task-specific requirements. In the following sections, we demonstrate how this capability allows our method to handle diverse scenarios ranging from \gls{hoi} to Self-Interaction planning via language cues.

\subsection{Human--Object Interaction}

A primary application of our method lies in \gls{hoi}. By synthesizing both body and hand (including finger) motions, our framework enables realistic grasping and manipulation. To demonstrate this, we generate stable grasps by employing GRABNet~\cite{GRAB}, a static grasp synthesis method that generates hand poses conditioned on the object geometry. GRABNet provides a static grasp configuration for a given object, which we assume remains stable throughout the object's motion. Under this assumption, we optimize the hand joint trajectories using a closeness loss \(\mathcal{L}_{\text{close}}\). This generates realistic full-body motions that interact with objects while having natural finger articulation before grasping the object. \cref{fig:qualitative_results} illustrates sample results, and additional visualizations are provided in the supplementary video. Our optimization framework is designed in a plug-and-play fashion, which allows seamless integration with various constraints. This flexibility enables our system to adapt to different object shapes and interaction scenarios, thereby facilitating diverse applications in Robotics and \gls{ar}/\gls{vr}.

\input{figures/tex_files/main_body/failure_cases}
\input{figures/tex_files/sup_mat/user_study_results}

\subsection{Self-Interaction using LLMs}

Our method can also be extended to scenarios involving Self-Interaction, where high-level semantic guidance is required. For this purpose we leverage recent advances in \gls{llm}s through a novel pipeline that generates structured contact plans. For this, we sample 236 vertices on the human body, each with a semantic label that may be important for the Self-Interaction task. These points, along with their semantic labels, are included in a textual prompt that also contains curated samples and explanations for meaningful interaction predictions. This prompt is given to an instance of GPT-o\(3\) without any fine-tuning~\cite{openai2024o3o4mini}. The model predicts which parts should be in contact and at what timestamps, based on a high-level action description. For instance, given prompts such as \textit{clap your hands}, the \gls{llm} outputs a set \( \nu = \{(j^{\tau_1}, k^{\tau_1}, \tau_1), (j^{\tau_2}, k^{\tau_2}, \tau_2), \ldots \} \) indicating the targeted vertices \( j^{\tau_i}, k^{\tau_i} \) and corresponding timestamps \( \tau_i \). This information takes into account the dynamics of the motion as it includes timestamps and is then enforced via a contact loss, defined as:

\[
\mathcal{L}_{\text{contact}}(\mathbf{x}, \nu) := \frac{1}{|\nu|} \sum_{(j^{\tau}, k^{\tau}, \tau) \in \nu}\left\|\hat{\mathbf{v}}_{j^\tau}^{}(\boldsymbol{x}) - \hat{\mathbf{v}}_{k^\tau}^{}(\boldsymbol{x})\right\|_1.
\]
Here \( \hat{\mathbf{v}}_{j^\tau}^{}(\boldsymbol{x}) \) indicates vertex location with index \( j^{\tau}\). This loss ensures that the specified body parts are in contact at the desired times, enabling structured and interpretable motion planning. \cref{fig:qualitative_results} shows the qualitative examples of self-contact guided by \gls{llm} outputs. For sampled points on the body and further qualitative examples, we refer to Sup.Mat.. 




\noindent\textbf{Failure Cases:} \methodname fails in highly dynamic scenarios (\cref{fig:failure_cases}), leading to self-penetration and unnatural motion.

%% file: figures/tex_files/main_body/qualitative_results.tex
\begin{figure*}
    \centerline{\includegraphics[clip, height=18.5cm]{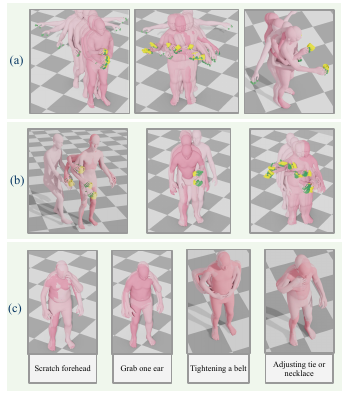}}
    \vspace{-0.5cm}
    \newcommand{\captitle}{\textbf{\methodname qualitative results}}
    \caption[\captitle]{\captitle: \methodname can track any given joint location (all \(10\) fingertips) over time (a). Conform a grasp provided by an off-the-shelf method(b) or Self-Interaction (c) In this case, it tracks the fingertips. Objects are provided only for visualization purposes, they do not play any role in the input. Mesh colors darken over time.}
    \label{fig:qualitative_results}
    \vspace{-0.3cm}
\end{figure*}

%% file: figures/tex_files/main_body/failure_cases.tex
\begin{figure}
    \centerline{\includegraphics[width=0.45\textwidth,clip]{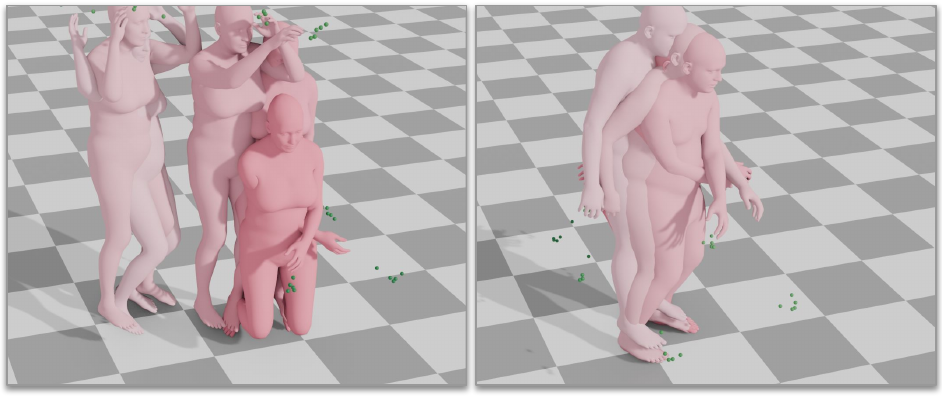}}
    \vspace{-0.2cm}
    \caption[]{\textbf{\methodname failure cases.} When the trajectory is highly dynamic, \methodname may fail to comply with task-specific requirements, which can lead to self-penetration and unrealistic motion. The example above illustrates this for the Keypoint Tracking task.}
    \label{fig:failure_cases}
    \vspace{-0.3cm}
\end{figure}

%% file: figures/tex_files/sup_mat/user_study_results.tex
\begin{figure}[h!]
    \centerline{\includegraphics[
    width=0.48\textwidth, clip]{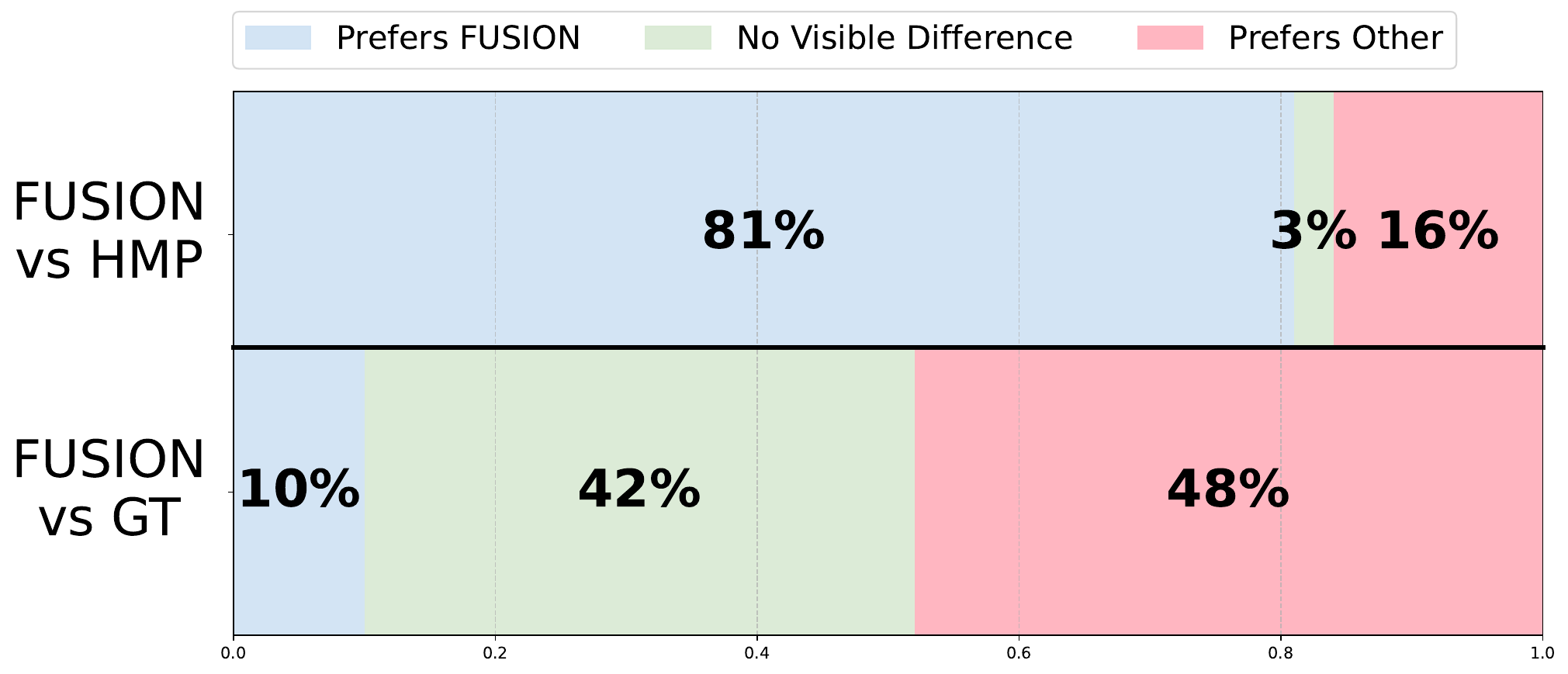}}
    \vspace{-0.2cm}
    \caption[]{\textbf{Subject preference ratios on test split.} Subjects were tasked with selecting one of three options: preferring \methodname, reporting no visible difference, and preferring the other method.}
    \label{fig:user_study_results}
    \vspace{-0.3cm}
\end{figure}

%% file: sections/5_experiments.tex
\section{Experiments}
\label{sec:experiments}

We provide experimental evaluations to test the performance of \methodname. We compare our approach with \Gls{sota} motion generation models on Keypoint Tracking performance and motion realism and perform ablation studies to assess design choices on motion synthesis quality.

\subsection{Comparison with \texorpdfstring{\Gls{sota}}{SOTA} methods}
\label{sec:evaluation}

We compare \methodname with previous \gls{sota} motion generation models: MDM, PriorMDM, GMD, OmniControl, and TLControl~\cite{MDM, PriorMDM, GMD, omnicontrol, TLControl}, on the Keypoint Tracking task using the HumanML3D dataset~\cite{HumanML3D}. We evaluate tracking performance on six key joints: Pelvis, Head, Left Wrist, Right Wrist, Left Foot, and Right Foot, using Trajectory Error, Location Error, and Average Error as per previous methods~\cite{TLControl, omnicontrol, MDM}. Notably, \methodname is unconditional and operates directly on a pose-based representation of the \gls{smpl-x} body model~\cite{SMPL-X:2019} instead of a skeletal representation of the SMPL model~\cite{SMPL:2015}. As such, unlike the other baselines, we cannot compute FID, KID, Diversity, and R-Precision metrics. \cref{tab:humanml3d_all} provides an overview of the performance on the HumanML3D dataset~\cite{HumanML3D}. This comparison demonstrates that our approach achieves competitive performance in terms of control accuracy and outperforms in motion realism~(\cref{sec:qual_comparison}). However, where our performance lags, it can be attributed to the trade-off between expressiveness and metric-specific evaluations that favor rigid skeletal representations.

\input{tables/sota_humanml3d}

\subsection{Qualitative Comparison}
\label{sec:qual_comparison}

Keypoint Tracking should be performed without compromising motion realism to preserve naturalness. Since all previous Keypoint Tracking methods operate on the SMPL body model, for a fair comparison, we evaluate the synthesized body and hand motions separately.

For comparing the realism of the synthesized body motions, we use a \gls{sota} method for motion evaluation: MotionCritic~\cite{motioncritic2025}. Given a body motion in SMPL pose format, MotionCritic evaluates the realism and outputs a numerical score. To ensure compatibility, we only use the body pose generated by \methodname (local hand motions excluded), with wrist orientations set to the identity rotation. As \cref{tab:humanml3d_all} shows, compared to the previous \gls{sota} joint tracking method TLControl~\cite{TLControl}, \methodname outperforms it on the motion realism metric. For hand motion synthesis, we conduct a perceptual study comparing \methodname with both the \gls{gt} and HMP~\cite{HMP} on the Keypoint Tracking task, which involves tracking all \(10\) fingertips on our test split. Participants were asked to evaluate the realism of motions involving the interaction of both hands with objects. \cref{fig:user_study_results} shows the subject preference ratios. The results indicate that \methodname synthesizes more realistic hand motions than HMP and achieves a level of realism comparable to the \gls{gt}. For more details, we refer to Sup.Mat..

\subsection{Quantitative Analysis of \texorpdfstring{\gls{llm}}{LLM}}
\label{sec:llm_quant}

Since high-level instructions for Self-Interaction are taken from an \gls{llm}, the quality of the synthesized motions depends on the quality of the high-level planning. To assess the reliability of the planning, we use a second instance of GPT-o\(3\)~\cite{openai2024o3o4mini} to perform the reverse reasoning, \ie, prompting it with only \(4\) output samples of the first instance of GPT-o\(3\) to get the instruction that describes the motion. Then, we compare the initial instructions with those obtained from the second model using BERTScore, focusing on the F1 metric, which provides a measure of semantic similarity~\cite{bert-score}. \cref{tab:llm_bert_scores_small} presents two representative sentence pairs with the highest and lowest F1 scores, illustrating the quality of alignment. These examples suggest that the model can successfully infer semantically aligned instructions from its outputs. For the details, we refer to Sup.Mat..

\input{tables/ablation}
\input{tables/bert_scores_small}

\subsection{Ablation Studies}
\label{sec:ablation }

\noindent\textbf{Unified Motion Prior Optimization:} To assess the benefits of a unified representation that jointly models hand and body motion, we conduct an ablation study comparing our method with a sequential optimization approach. In the sequential variant, the model is first optimized for hand motion (given 10 fingertip trajectories) and subsequently for body motion (given wrist trajectories from optimized hand motions). Results are shown in \cref{tab:ablation}. The first row refers to our baseline method. The second row stands for the sequential variant of our method. Our experiments indicate that the sequential approach leads to increased foot skating and higher trajectory errors. In contrast, optimizing the unified motion prior not only improves performance but also offers superior computational efficiency. To ensure a fair comparison, all experiments used identical learning rates, loss functions, and epoch counts. We refer to Sup.Mat..

\noindent\textbf{Comparison with combined off-the-shelf priors:} Since to the best of our knowledge, no previous motion-prior includes both body and hands in a unified model, we evaluate our method against a combined model that integrates \gls{sota} hand and body motion priors (HMP and TLControl, respectively)~\cite{HMP, TLControl}. In this baseline, hand motion is first optimized, and the resulting wrist trajectories are used to subsequently optimize the body motion. We report the results in the last row of \cref{tab:ablation} and show that \methodname outperforms this two-stage process in overall tracking quality metrics.

%% file: tables/sota_humanml3d.tex
\begin{table*}[htbp]
\centering
\resizebox{0.95\textwidth}{!}{
\begin{tabular}{c|c|c|c|c|c|c|c|c|c|}
\hline
Method & \makecell[c]{Control \\Joint} & \makecell[c]{Traj. Error \(\downarrow\)\\(50 cm, \%)} & \makecell[c]{Traj. Error \(\downarrow\)\\(10 cm, \%)} & \makecell[c]{Traj. Error \(\downarrow\)\\(5 cm, \%)} & \makecell[c]{Loc. Error \(\downarrow\)\\(50 cm, \%)} & \makecell[c]{Loc. Error \(\downarrow\)\\(10 cm, \%)} & \makecell[c]{Loc. Error \(\downarrow\)\\(5 cm, \%)} & \makecell[c]{Avg. Error \(\downarrow\)\\(cm)} & \makecell[c]{Motion \\ Realism \(\uparrow\) \cite{motioncritic2025}} \\
\hline
MDM & & 40.22 & - & - & 30.76 & - & - & 59.59 & - \\
PriorMDM & & 34.57 & - & - & 21.32 & - & - & 44.17 & - \\
GMD & Pelvis & 9.31 & - & - & 3.21 & - & - & 14.39 & - \\
OmniControl & & 3.87 & - & - & 0.96 & - & - & 3.38 & - \\
TLControl & & \first{0.00} & \first{0.39} & \first{11.04} & \first{0.00} & \first{0.00} & \first{0.26} & \first{1.08} & -3.89 \\
\textbf{Ours} &  & \first{0.00} & 1.47 & 12.41 & \first{0.00} & 0.13 & 0.68 & 1.16 & \first{0.37} \\
\hline
OmniControl & & 4.22 & - & - & 0.79 & - & - & 3.49 & - \\
TLControl & Head & \first{0.00} & \first{1.10} & 22.95 & \first{0.00} & \first{0.07} & 0.89 & \first{1.10} & -7.47 \\
\textbf{Ours} & & \first{0.00} & 2.03 & \first{13.24} & \first{0.00} & 0.15 & \first{0.44} & 1.51 & \first{-0.32} \\
\hline
OmniControl & & 8.01 & - & - & 1.34 & - & - & 5.29 & - \\
TLControl & Left Wrist & \first{0.00} & \first{1.37} & 12.88 & \first{0.00} & \first{0.11} & \first{0.28} & \first{1.08} & -5.66 \\
\textbf{Ours} & & \first{0.00} & 2.93 & \first{5.54} & \first{0.00} & 0.13 & 0.35 & 1.23 & \first{-0.79} \\
\hline
OmniControl & & 8.13 & - & - & 1.27 & - & - & 5.19 & - \\
TLControl & Right Wrist & \first{0.00} & \first{1.17} & 12.30 & \first{0.00} & \first{0.02} & \first{0.18} & \first{1.09} & -5.62 \\
\textbf{Ours} & & \first{0.00} & 3.03 & \first{6.57} & \first{0.00} & 0.15 & 0.38 & 1.31 & \first{-0.83} \\
\hline
OmniControl & & 5.94 & - & - & 0.94 & - & - & 3.14 & - \\
TLControl & Left Foot & \first{0.00} & \first{0.00} & \first{10.12}  & \first{0.00} & \first{0.01} & \first{0.15} & \first{1.14} & -4.79 \\
\textbf{Ours} & & \first{0.00} & 3.62 & 13.04 & \first{0.00} & 0.54 & 0.87 & 1.42 & \first{-1.13} \\
\hline
OmniControl & & 6.66 & - & - & 1.20 & - & - & 3.34 & - \\
TLControl & Right Foot & \first{0.00} & \first{0.49} & \first{11.13} & \first{0.00} & \first{0.00} & \first{0.16} & \first{1.16} & -3.15 \\
\textbf{Ours} & & \first{0.00} & 2.75 & 12.82 & \first{0.00} & 0.24 & 0.74 & 1.46 & \first{-0.72} \\
\hline
OmniControl & & 75.59 & - & - & 12.30 & - & - & 23.67 & - \\
TLControl & All Joints above & \first{0.00} & \first{10.62} & 71.09 & \first{0.00} & \first{0.09} & \first{1.71} & \first{1.57} & -7.86 \\
\textbf{Ours} & & \first{0.00} & 35.28 & 
\first{64.14} & \first{0.00} & 1.05 & 3.73 & 1.78 & \first{-1.26} \\
\hline
\end{tabular}
}
\vspace{-0.3cm}
\caption[]{Quantitative comparison with previous \gls{sota} methods. The best scores are highlighted bold for Keypoint Tracking on HumanML3D. Although \methodname controls pose rather than spatial joint locations, it performs on par with \gls{sota} skeletal control methods. On motion realism our method outperforms previous \gls{sota} method TLControl \cite{TLControl} and reports close numbers to the ground truth.}
\label{tab:humanml3d_all}
\vspace{-0.25cm}
\end{table*}

%% file: tables/ablation.tex
\begin{table*}[h]
\centering
\resizebox{0.95\textwidth}{!}{%
\begin{tabular}{c|c|c|c|c|c|c}
\hline
Method & Control Joint & \shortstack{Skating \\ ratio (\%) $\downarrow$} & \shortstack{Loc. Err. $\downarrow$ \\ (30 cm, \%)} & \shortstack{Loc. Err. $\downarrow$ \\ (10 cm, \%)} & \shortstack{Loc. Err. $\downarrow$ \\ (5 cm, \%)} & \shortstack{Avg. Err. \\ (cm) $\downarrow$} \\
\hline
\methodname & Finger Tips & \first{6.03} & \first{0.00} & \first{0.03} & \first{0.29} & \first{0.59} \\ \hline
\shortstack{Sequential \methodname}  & \shortstack{Finger Tips + Wrists} & 13.80 & \first{0.00} & 0.25 & 2.19 & 1.23 \\ \hline
\shortstack{HMP~\cite{HMP} + TLControl~\cite{TLControl}} & \shortstack{Finger Tips + Wrists} & 15.35 & \first{0.00} & 0.32 & 4.02 & 1.64 \\
\hline
\end{tabular}
}
\vspace{-0.2cm}
\caption{Ablation studies on the test split. Results emphasize the advantage of having a compact method over optimizing two separate priors. In addition, \methodname outperforms the combination of \gls{sota} hand and body motion models in Keypoint Tracking.}
\label{tab:ablation}
\vspace{-0.3cm}
\end{table*}


%% file: tables/bert_scores_small.tex
\begin{table}[htbp]
\centering
\renewcommand{\arraystretch}{1.5}

\resizebox{0.45\textwidth}{!}{%
\begin{tabular}{c|>{\centering\arraybackslash}m{5cm}|c|c|c}
\hline
Instruction & Predicted Sentence & R & P & F1 \\
\hline
\makecell[c]{Stay in  \\ attention position} &
\makecell[c]{Stand with both palms on your \\ back thighs, heels pressed together.} & 0.804 & 0.830 & 0.817 \\ 
\hline
\makecell[c]{Put your hands on your hips} & \makecell[c]{Place your hands on your hips.} & 0.981 & 0.977 & 0.979 \\
\hline
\multicolumn{2}{c|}{Average values} &  0.894 & 0.9 & 0.897 \\
\hline
\end{tabular}
} 
\vspace{-0.3cm}
\caption[]{Examples of instruction–\gls{llm} prediction pairs for the highest, lowest, and average BERTScore F1 values.}
\label{tab:llm_bert_scores_small}
\vspace{-0.45cm}
\end{table}

%% file: sections/6_conclusion.tex
\section{Conclusion}
\label{sec:conclusion}
 
In this work, we present \methodname, the first unconditional full-body motion prior on SMPL-X, jointly modeling both body and articulated hand motions. Unlike existing approaches that treat hands and body separately, our method integrates heterogeneous datasets and leverages a denoising diffusion model to capture fine-grained motion dynamics across the entire body. We further adapt a flexible optimization framework that refines the latent diffusion noise under task-specific constraints. Experiments show that \methodname achieves competitive performance with the skeleton-based \gls{sota} method in Keypoint Tracking and surpasses it in terms of motion realism. Beyond conventional uses of motion priors, \methodname enables applications such as \gls{hoi} and Self-Interaction. Overall, \methodname provides a promising data-driven solution for realistic full-body motion synthesis in fields such as Computer Vision, Robotics, and \gls{ar}/\gls{vr}. 


%% file: sections/supmat.tex
\clearpage
\setcounter{page}{1}
\maketitlesupplementary

\section{Architecture}
\label{sec:architecture}

\Cref{fig:architecture} shows the model architecture of our motion prior. It is a transformer-based diffusion model with \(19.7\)M trainable parameters.

\section{Optimization Hyperparameters}
\label{sec:optimization_supmat}

\cref{tab:opt_configuration} lists the optimization hyperparameters used in evaluation and ablation studies. The chosen configuration balances task compliance (\(\lambda_{\text{close}}\)) with the plausibility of the resulting motion (\(\lambda_{\text{foot}}\), \(\lambda_{\text{ch}}\), \(\lambda_{\text{lk}}\)).

\input{tables/optimization_config}

\input{figures/tex_files/sup_mat/architecture}

\section{Evaluation Metrics}
\label{sec:supmat_eval_metrics}

Trajectory Error is the percentage of unsuccessful trajectories, defined as those with any keyframe location error exceeding a specified distance threshold. Location Error is the percentage of keyframe locations that exceed this threshold. Average Error measures the mean distance between the generated motion locations and the keyframe locations, measured at the keyframe motion steps. 

\section{Dataset Curation}
\label{sec:supmat_dataset_curation}

\cref{fig:smplx_mano_smplx} shows an example of random merging of the global body motion with local hand motion. Although it is rare, random merging of the local hand motion and the global body motion might introduce additional self-intersection. To address this issue, we perform a self-penetration test to ensure the obtained motions are physically possible. We compute face intersections between the hand mesh and the rest of the body mesh caused by copying the local hand motion. If there is an intersection exceeding a threshold, we discard this random merging in favor of another randomly sampled local hand motion. 

\input{figures/tex_files/sup_mat/dataset_ratios_body}
\input{figures/tex_files/sup_mat/dataset_ratios_hand}
\input{figures/tex_files/sup_mat/body_hand_merging} 

\section{Qualitative Results}
\label{sec:supmat_qualitative_results}

For hand motion synthesis, we conducted a perceptual study comparing \methodname with both the \gls{gt} and HMP~\cite{HMP} on the Keypoint Tracking task, which involves tracking all \(10\) fingertips on our test split. Participants were asked to evaluate the realism of motions involving the interaction of both hands with objects. Videos were presented side-by-side in random order as shown in \cref{fig:user_study_gui}. Objects were included for visualization purposes only, to help users better distinguish between the motions. The evaluation criteria included hand jitter, naturalness of articulation, and object penetration. Subject preference ratios from \(3\) participants indicate that \methodname synthesizes more realistic hand motions than HMP and achieves a level of realism comparable to the \acrlong{gt}. \cref{fig:user_study_gui} shows the user panel.

\input{figures/tex_files/sup_mat/user_study_gui}

\section{Self-Interaction}
\label{sec:supmat_self_interaction}

We use our optimization framework with a prompted instance of GPT-o\(3\)~\cite{openai2024o3o4mini} to curate the self-interaction dataset. In total, there are 94 sequences with textual descriptions, contact conditions, and the corresponding human motions after optimization. This small dataset will be shared in the codebase. 

\subsection{Sampling Body Vertices}

We sample 236 vertices along with their semantic labels (\eg \textit{right\_elbow\_inside, left\_biceps}) on the human body that may be important for self-interaction tasks. \Cref{fig:self_interaction_body_vertices,fig:self_interaction_hand_vertices,fig:self_interaction_head_vertices} show the sampled vertices with a focus on the body, hands, and head, respectively. 

\input{figures/tex_files/sup_mat/self_interaction_body_vertices}

\input{figures/tex_files/sup_mat/self_interaction_hand_vertices}
\input{figures/tex_files/sup_mat/self_interaction_head_vertices}

\subsection{Quantitative Evaluation}

\input{figures/tex_files/sup_mat/bert_scores_pipeline}

\cref{fig:bert_scores_pipeline} shows the devised pipeline for the evaluation of \gls{llm} planning. We run this all \(94\) every input sentence in our self-interaction. \cref{tab:bert_scores_full} shows every input--predicted sentence pair and the corresponding BERTScore values~\cite{bert-score}. The last two rows are for the pairs with the highest and the lowest F1 score. 

The BERTScore analysis of the input–predicted sentence pairs shows generally high alignment between the intended actions and their corresponding descriptions, with most F1 scores clustering above \(0.85\), indicating strong semantic similarity. Actions involving direct tactile interactions, such as \textit{rubbing the palms together} (F1 = \(0.953\)), \textit{rub your left calf} (F1 = \(0.962\)), and \textit{smell your armpit} (F1 = \(0.977\)), achieve particularly high scores, reflecting clear and unambiguous mappings between input and evaluation sentences. Gestures that are slightly more abstract or involve sequential motion, such as \textit{do a squat} (F1 = \(0.932\)) or \textit{crossing arms} (F1 = \(0.873\)), show slightly lower but still reasonable alignment. Overall, the table reflects that the predictions are semantically faithful to the inputs, with minor variations in descriptive richness or specificity influencing BERTScore variations.

\input{tables/bert_scores_full}

%% file: tables/optimization_config.tex
\begin{table}[htbp]
\centering
\resizebox{0.4\textwidth}{!}{
\begin{tabular}{c|c|c|c|c|c}
Stage & Epoch & \(\lambda_{\text{lk}}\) & \(\lambda_{\text{foot}}\) & \(\lambda_{\text{ch}}\) & \(\lambda_{\text{close}}\)  \\
\hline
1 & 800 & 0.5 & 0.5 & 0.5 & 1 \\ 
2 & 800 & 0.1 & 0.1 & 0.1 & 1 \\ 
\hline
\end{tabular}
}
\caption{Optimization configuration. The same configuration applies across all ablation and evaluation settings.}
\vspace{-0.5cm}
\label{tab:opt_configuration}
\end{table}

%% file: figures/tex_files/sup_mat/architecture.tex
\begin{figure*}[t]
\centerline{\includegraphics[width=0.95\textwidth, clip]{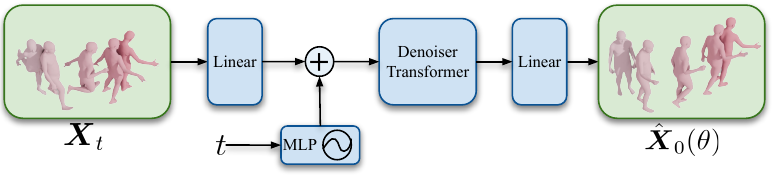}}
    \newcommand{\captitle}{\textbf{\methodname Model Architecture}}
    \caption[\captitle]{\captitle: Given a noisy motion \(\boldsymbol{X}_t\) and the corresponding noising timestep \(t\), \methodname first performs linear projection, and concatenates it with the time embeddings of the noising timestep. This concatenation forms the input to the Denoiser Transformer. Then it performs linear projection to the Denoiser Transformer output and obtains denoised motion \(\hat{\boldsymbol{X}}_0(\theta)\).}
    \label{fig:architecture}
\end{figure*}

%% file: figures/tex_files/sup_mat/dataset_ratios_body.tex
\begin{figure}[htbp]
    \centerline{\includegraphics[width=0.45\textwidth,clip]{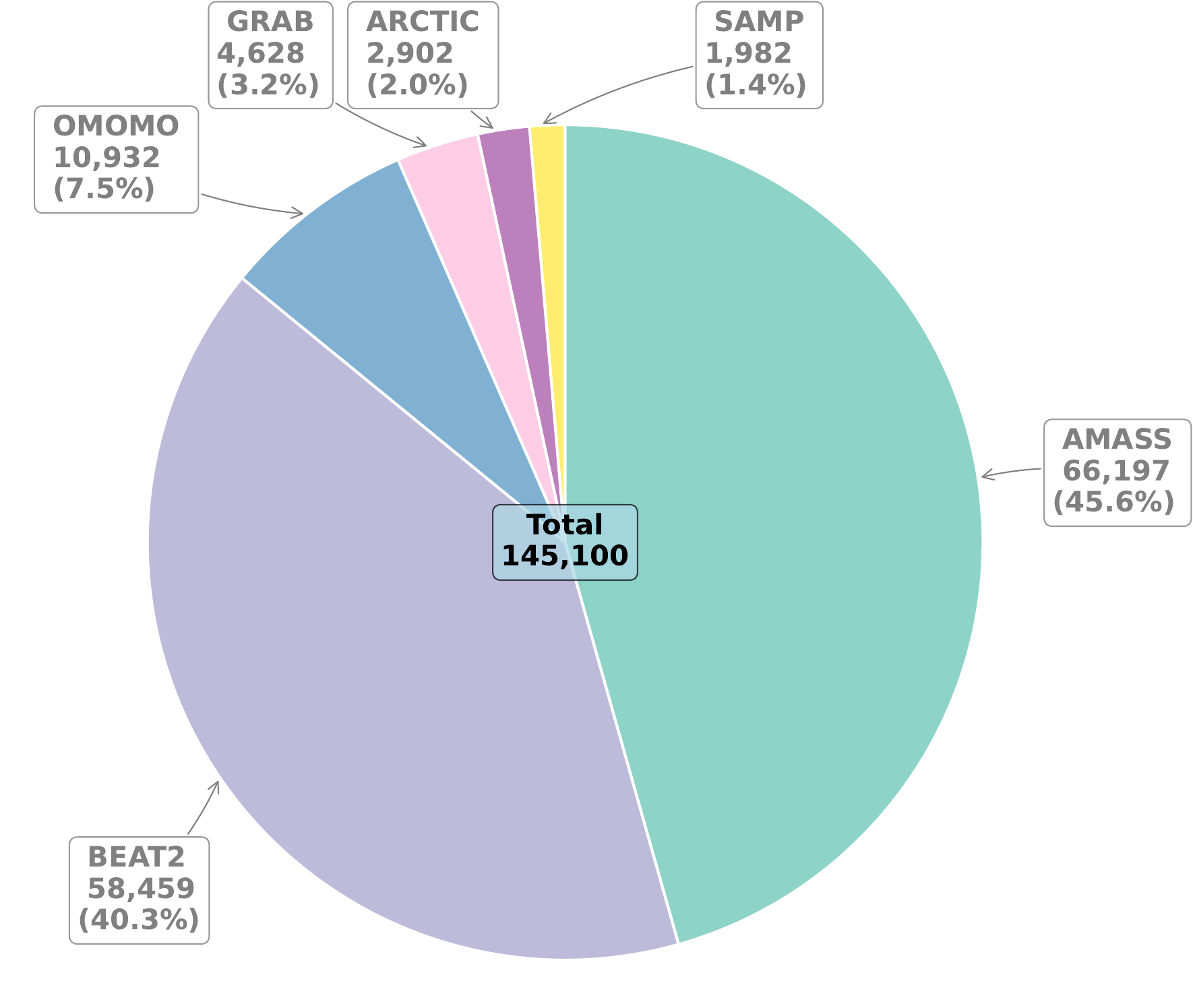}}
    \newcommand{\captitle}{\textbf{Body Dataset Sequences}}
    \caption[\captitle]{\captitle: Distribution of sequences across body motion datasets used for training. AMASS and BEAT2 constitute the majority of the training corpus, while smaller specialized datasets (OMOMO, GRAB, ARCTIC, and SAMP) provide additional diversity to the overall collection.}
    \label{fig:body_dataset_ratios}
\end{figure}

%% file: figures/tex_files/sup_mat/dataset_ratios_hand.tex
\begin{figure}
    \centerline{\includegraphics[width=0.45\textwidth,clip]{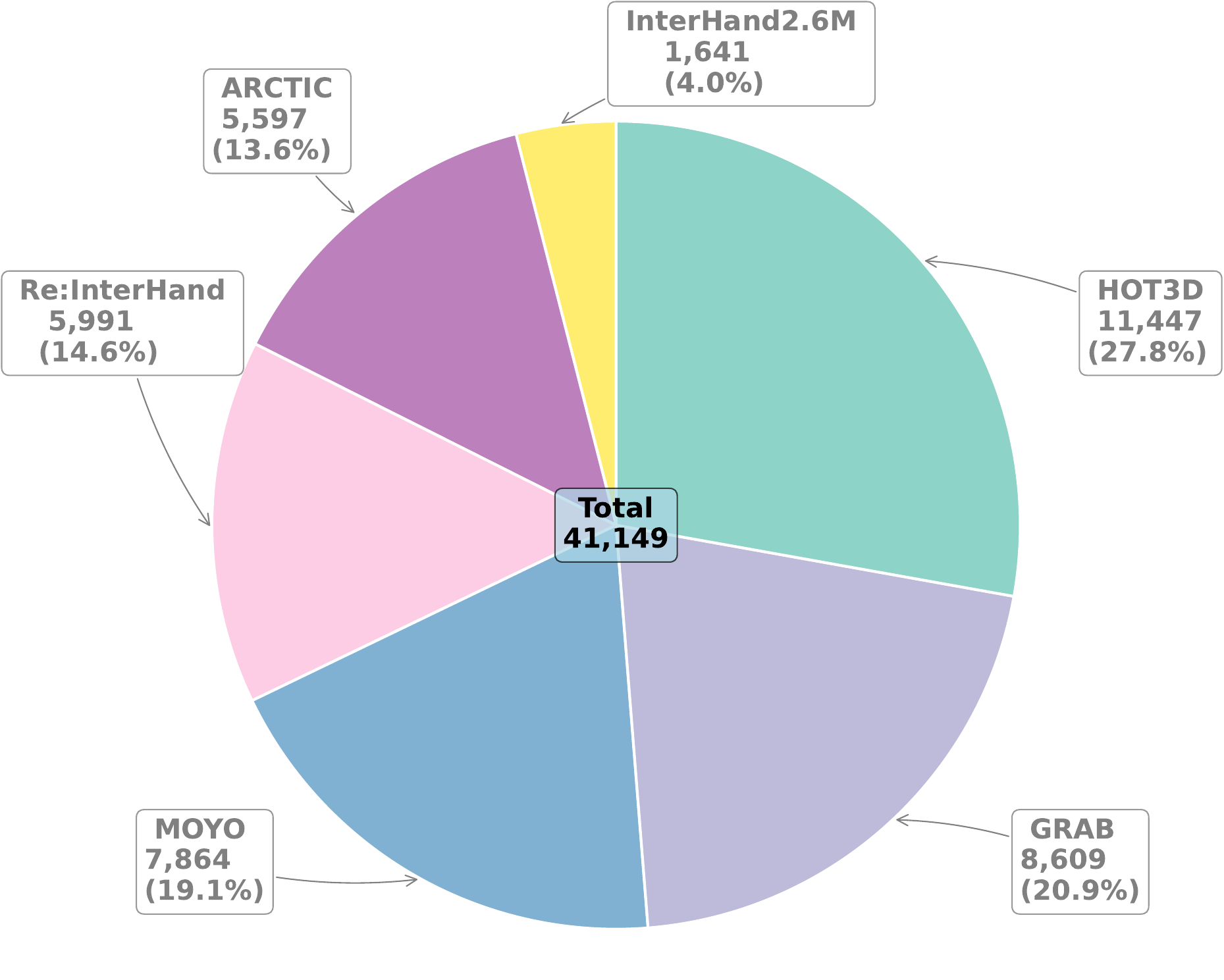}}
    \newcommand{\captitle}{\textbf{Hand Dataset Sequences}}
    \caption[\captitle]{\captitle: Distribution of sequences across hand motion datasets used for training. HOT3D constitutes the majority of the training corpus. GRAB, MOYO, Re:InterHand, ARCTIC, and InterHand2.6M datasets provide additional diversity to the overall collection.}
    \label{fig:hand_dataset_ratios}
\end{figure}

%% file: figures/tex_files/sup_mat/body_hand_merging.tex
\begin{figure*}[ht]
    \centerline{\includegraphics[width=0.9\textwidth,clip]{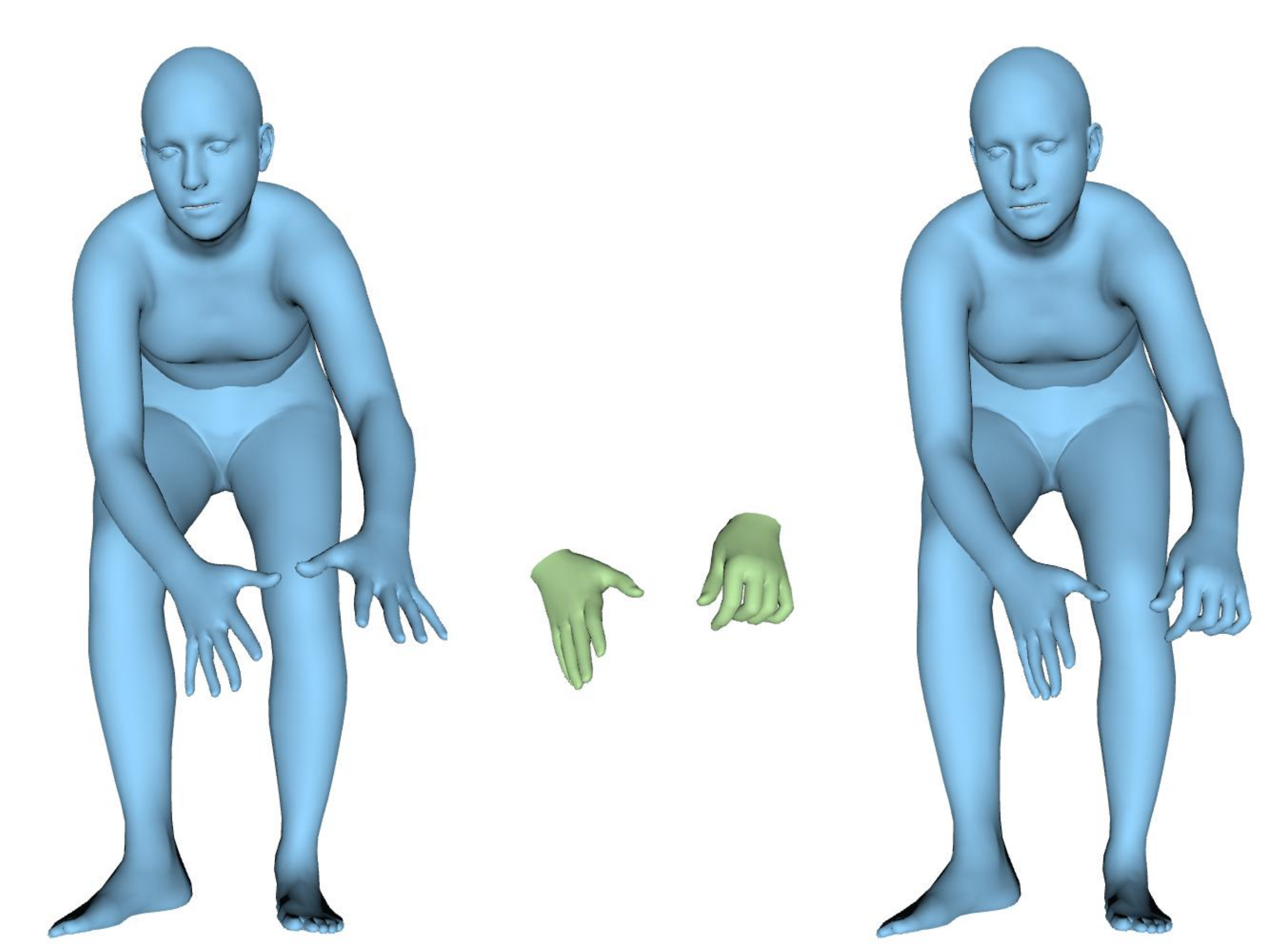}}
    \newcommand{\captitle}{\textbf{\methodname Visualization of Random Body--Hand Pairing}}
    \caption[\captitle]{\captitle: The mesh from the body dataset is shown on the left. The middle panel presents local hand meshes with randomly sampled poses, but with wrist orientations and locations taken from the \gls{gt} data. On the right, we show the resulting full-body mesh obtained by combining the sampled hand poses with the corresponding body pose. Notably, self-penetration is effectively prevented, since the wrist orientations are directly derived from the body, ensuring consistent and plausible articulation.}
    \label{fig:smplx_mano_smplx}
\end{figure*}

%% file: figures/tex_files/sup_mat/user_study_gui.tex
\begin{figure*}[!ht]
    \centerline{\includegraphics[
    width=0.9\textwidth, clip]{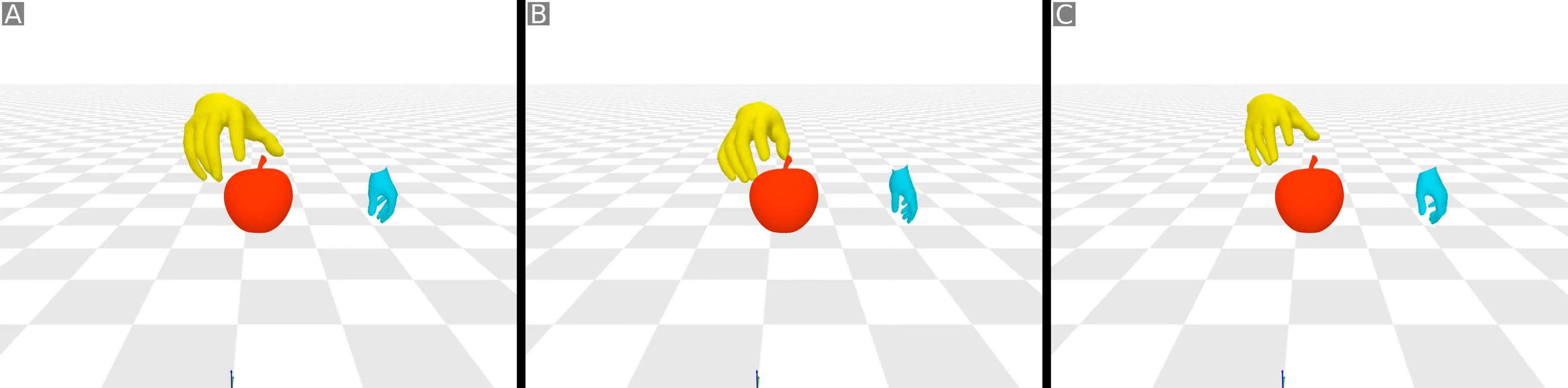}}
    \newcommand{\captitle}{\textbf{User Study Graphical Interface}}
    \caption[\captitle]{\captitle: Subjects were tasked with ranking the realism of the three options A, B, and C. Ordering is random for each interaction sample.}
    \label{fig:user_study_gui}
\end{figure*}

%% file: figures/tex_files/sup_mat/self_interaction_body_vertices.tex
\begin{figure*}
    \centerline{\includegraphics[width=0.9\textwidth, clip]{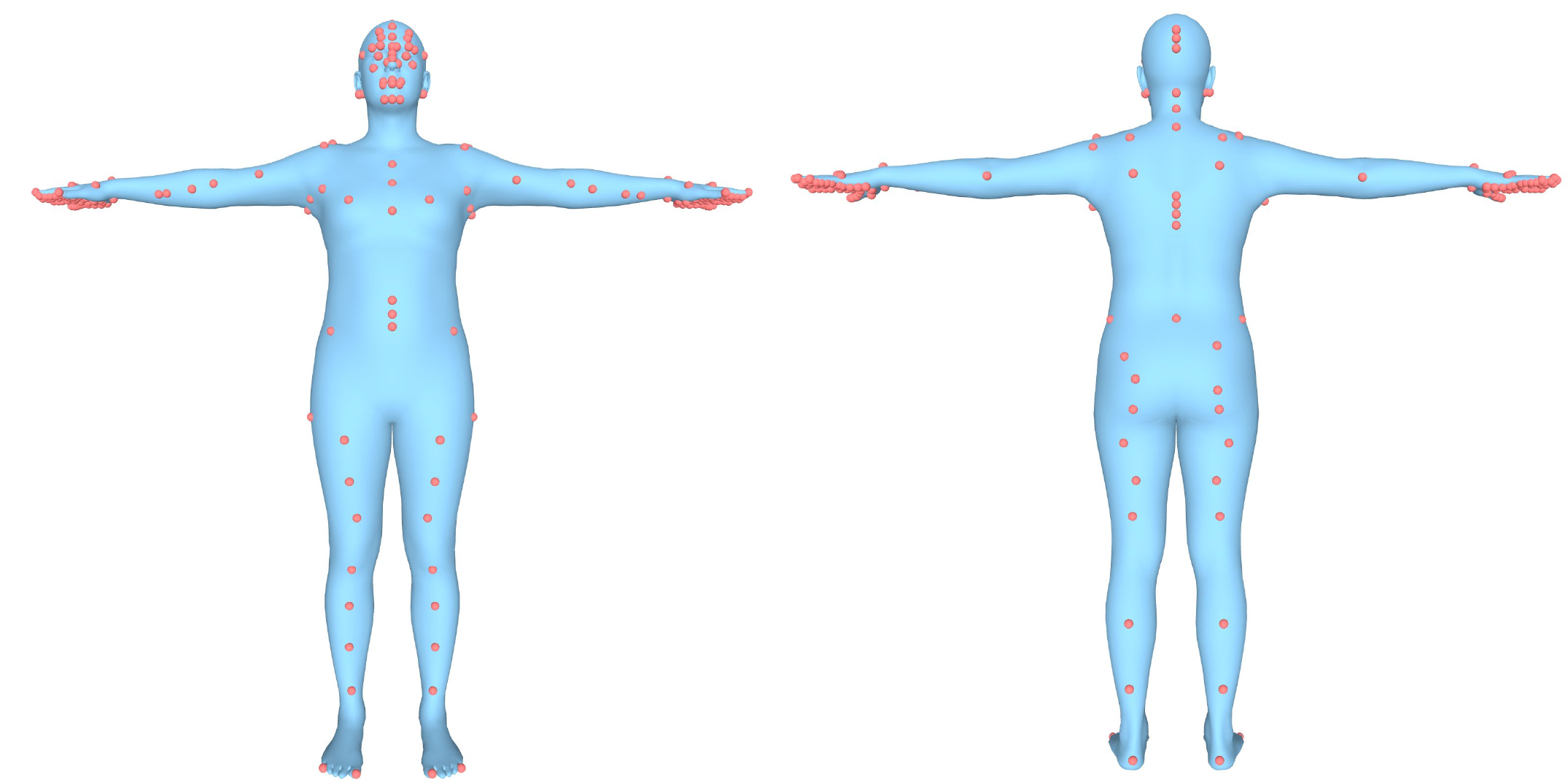}}
    \newcommand{\captitle}{\textbf{Sampled vertices for the Self-Interaction Task}}
    \caption[\captitle]{\captitle: We manually select \(236\) body vertices together with their semantic labels (\eg \textit{right\_fingertip}, \textit{left\_biceps}, \ldots). The selected vertices are distributed across major body parts but are denser in the hands, as fine-grained hand articulations are more likely to come into contact during interaction. This manual selection ensures that the set covers contact-relevant regions, while keeping the number of vertices computationally manageable. }
    \label{fig:self_interaction_body_vertices}
\end{figure*}


%% file: figures/tex_files/sup_mat/self_interaction_hand_vertices.tex
\begin{figure*}
    \centerline{\includegraphics[width=0.9\textwidth, clip]{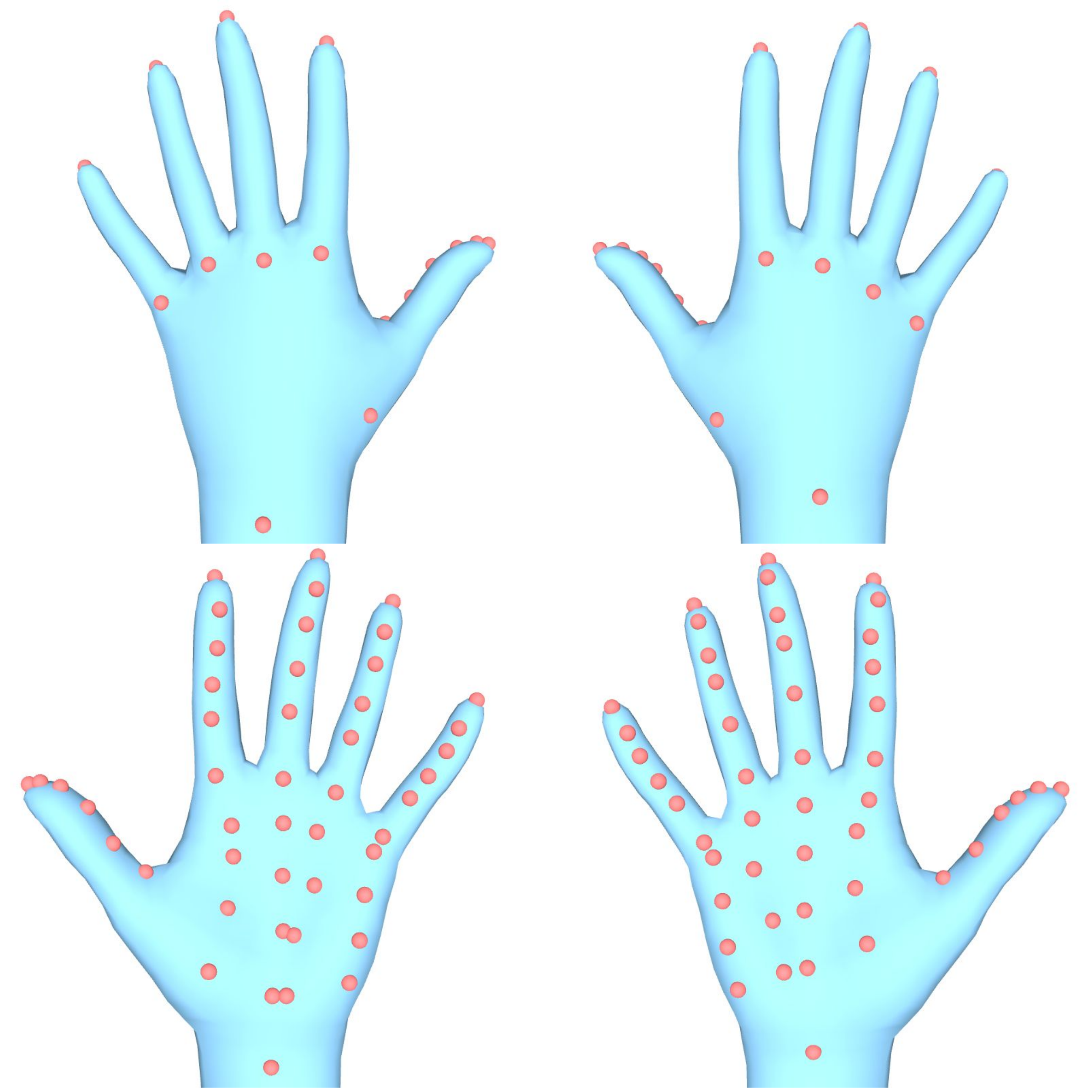}}
    \newcommand{\captitle}{\textbf{Sampled hand vertices for the Self-Interaction Task}}
    \caption[\captitle]{\captitle: The palms contain more sampled points than the back of hands to better capture contact in relevant tasks.}
    \label{fig:self_interaction_hand_vertices}
\end{figure*}

%% file: figures/tex_files/sup_mat/self_interaction_head_vertices.tex
\begin{figure*}
    \centerline{\includegraphics[width=0.9\textwidth, clip]{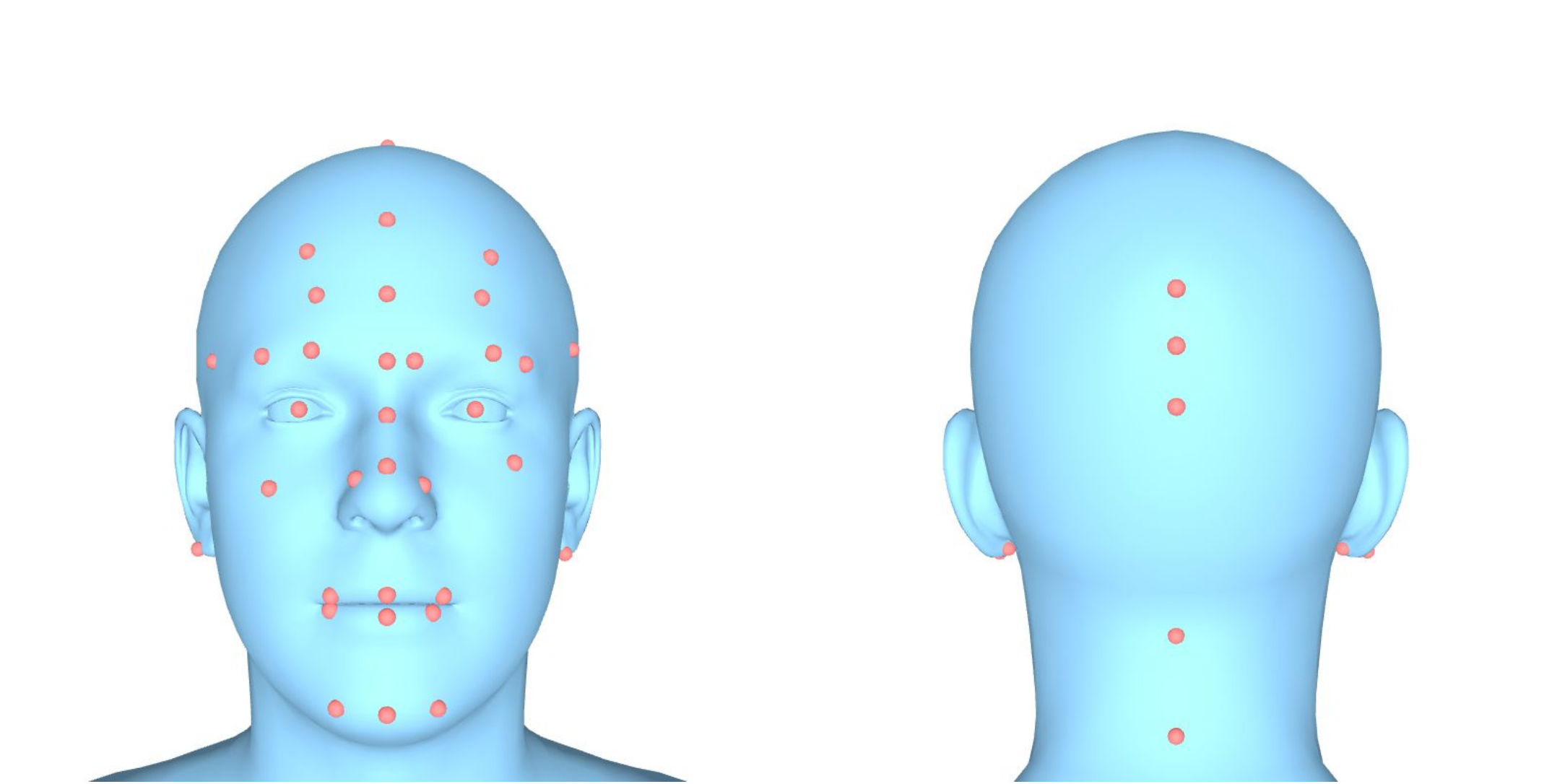}}
    \newcommand{\captitle}{\textbf{Sampled head vertices for the Self-Interaction Task}}
    \caption[\captitle]{\captitle: The face contains more sampled points than the back of the head to better capture contact in tasks such as touching the face.}
    \label{fig:self_interaction_head_vertices}
\end{figure*}

%% file: figures/tex_files/sup_mat/bert_scores_pipeline.tex
\begin{figure*}
    \centerline{\includegraphics[width=0.95\textwidth, clip]{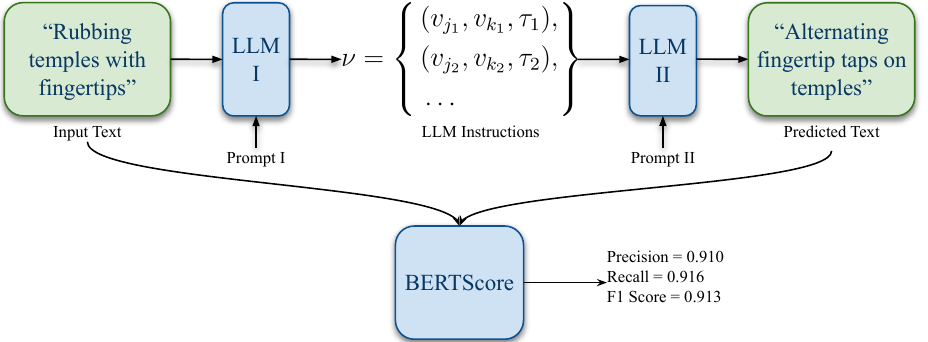}}

    \newcommand{\captitle}{\textbf{Quantitative Analysis of \gls{llm} Outputs}}
    \caption[\captitle]{\captitle: We employ a pipeline to verify the quality of \gls{llm}-generated plans. Specifically, we prompt a second instance of GPT-o3 with only four samples from the interaction dataset to generate a sentence describing the instruction. The second instance is then used in tandem with the first one to perform reverse reasoning. Then for each resulting input–prediction sentence pair, BERT is used to compute precision, recall, and F1 score based on semantic similarity~\cite{bert-score}.}
    \label{fig:bert_scores_pipeline}
\end{figure*}

%% file: tables/bert_scores_full.tex
\onecolumn 
    
    \newcommand{\captitle}{\textbf{BERTScores for input--predicted sentence pairs}}
    \renewcommand{\arraystretch}{1.4}
    \footnotesize
    \setlength{\LTpre}{0pt}
    \setlength{\LTpost}{0pt}
    \begin{longtable}{>{\raggedright\arraybackslash}p{0.8cm} >{\raggedright\arraybackslash}p{4.5cm} >{\raggedright\arraybackslash}p{5.5cm} >{\centering\arraybackslash}p{1.3cm} >{\centering\arraybackslash}p{1.3cm} >{\centering\arraybackslash}p{1.3cm}}
    \toprule
    \textbf{Index} & \textbf{Input Sentence} & \textbf{Evaluation Sentence} & \textbf{Precision} & \textbf{Recall} & \textbf{F1} \\
    \midrule
    \endfirsthead

    \toprule
    \textbf{Index} & \textbf{Input Sentence} & \textbf{Evaluation Sentence} & \textbf{Precision} & \textbf{Recall} & \textbf{F1} \\
    \midrule
    \endhead

    \bottomrule
    \endfoot

1 & Interlacing fingers & Press both palms and matching fingertips together in a prayer gesture. & 0.862 & 0.852 & 0.857 \\
2 & Cracking knuckles & Slide your left thumb across the right knuckles, then press palms together. & 0.842 & 0.859 & 0.85 \\
3 & Self fist bumping & Tap the knuckles of your hands together. & 0.868 & 0.852 & 0.86 \\
4 & Rubbing the palms together & Rub your palms together. & 0.954 & 0.952 & 0.953 \\
5 & Rubbing fingertips together & Alternately tap your left thumb with your index and middle fingertips. & 0.854 & 0.885 & 0.869 \\
6 & Pressing thumb against palm & Slide your right thumb across your palm. & 0.906 & 0.908 & 0.907 \\
7 & Rolling a ring around a finger & Rub your right ring finger with your thumb and middle fingertip. & 0.862 & 0.9 & 0.881 \\
8 & Flicking one finger with another & Right-hand index taps between thumb and middle finger. & 0.876 & 0.866 & 0.871 \\
9 & Massaging palm with opposite hand & Rub the left palm with the right thumb and index finger. & 0.892 & 0.902 & 0.897 \\
10 & Tapping fingers against each other & Bring all corresponding fingertips of both hands together in a prayer-like touch. & 0.86 & 0.875 & 0.867 \\
11 & Snap your fingers to make some noise & Snap your fingers with your right hand. & 0.937 & 0.916 & 0.926 \\
12 & Pulling fingers one by one with the opposite hand & Right hand pinches each left fingertip in sequence. & 0.884 & 0.891 & 0.887 \\
13 & Drumming left fingers on the back of the right hand & Tap your left fingertips on the knuckles of your right hand. & 0.919 & 0.908 & 0.914 \\
14 & Tracing a circle or shape on the palm with a finger & Run your right index finger across your left palm. & 0.889 & 0.881 & 0.885 \\
15 & Scratching forehead & Slide your left index finger across your forehead. & 0.853 & 0.851 & 0.852 \\
16 & Rub your back of head & Scratch the back of your head with the left hand. & 0.91 & 0.922 & 0.916 \\
17 & Tapping cheek with fingers & Touch your right cheek with your right fingertips. & 0.896 & 0.911 & 0.903 \\
18 & Scratching chin or jawline & Thoughtful chin stroking. & 0.88 & 0.86 & 0.87 \\
19 & Wiping sweat from forehead & Wipe your forehead with your right middle knuckle. & 0.88 & 0.922 & 0.901 \\
20 & Resting fingertips against forehead & Touch your forehead with right fingertips. & 0.918 & 0.921 & 0.92 \\
21 & Gently pinching and releasing fingers together & Do the OK gesture with the right hand. & 0.883 & 0.854 & 0.868 \\
22 & Fixing hair & Slide right fingertips from temple over head to neck. & 0.845 & 0.873 & 0.859 \\
23 & Touch you ear & Gently pinch your left earlobe. & 0.881 & 0.907 & 0.894 \\
24 & Grab your ear & Gently pinch your right earlobe with your right hand. & 0.867 & 0.924 & 0.894 \\
25 & Cover both ears & Massage your earlobes with both palms. & 0.881 & 0.912 & 0.896 \\
26 & Pinching own skin & Pinch your left forearm with your right thumb and index finger. & 0.82 & 0.856 & 0.837 \\
27 & Adjusting glasses & Right hand pinches nose then earlobe. & 0.859 & 0.862 & 0.861 \\
28 & Rubbing your knee & Stroke your left knee with your left hand. & 0.871 & 0.904 & 0.887 \\
29 & Rub your left calf & Rub your left calf with your left hand. & 0.945 & 0.98 & 0.962 \\
30 & Clapping hands twice & Rub your hands together. & 0.921 & 0.893 & 0.907 \\
31 & Rubbing hands together & Rubbing palms together. & 0.978 & 0.972 & 0.975 \\
32 & Grab your opposite ear & Pinch your right earlobe with the left hand. & 0.891 & 0.921 & 0.905 \\
33 & Put your hands on your hips & Place your hands on your hips. & 0.981 & 0.977 & 0.979 \\
34 & Both hands grabbing earlobes & Gently pinch both earlobes with thumb and index fingers. & 0.882 & 0.912 & 0.897 \\
35 & Place your hands on your waists & Stand with hands on hips. & 0.925 & 0.886 & 0.905 \\
36 & Fixing sleeves or rolling them up & Slide your right fingertips up your left arm. & 0.861 & 0.86 & 0.86 \\
37 & Hands on knees to rest after running & Rest both palms on your knees. & 0.898 & 0.882 & 0.89 \\
38 & Clean your armpit with your left hand & Scratch your left armpit with your left hand. & 0.958 & 0.965 & 0.961 \\
39 & Scretch yourself by touching your toes & Slide your hands down your legs to touch your toes. & 0.898 & 0.876 & 0.887 \\
40 & Touch your left heel with the left hand & Rub your left heel with your left hand. & 0.971 & 0.972 & 0.971 \\
41 & Crossing arms and tapping arm with fingers & Lightly tap both biceps with opposite fingertips. & 0.872 & 0.866 & 0.869 \\
42 & Placing both hands in front of the abdomen & Rest both hands on your belly. & 0.935 & 0.919 & 0.927 \\
43 & Wrapping fingers around the opposite wrist & Hold your right wrist with your left hand. & 0.887 & 0.902 & 0.894 \\
44 & Touching the back of the neck with one hand & Gently massage your neck with the left hand. & 0.927 & 0.919 & 0.923 \\
45 & Secure your head in the moment of emergency & Scalp massage. & 0.873 & 0.871 & 0.872 \\
46 & Pressing fingers against biceps or shoulders & Gently scratching left upper arm. & 0.888 & 0.866 & 0.877 \\
47 & Scratching forearm or wrist with opposite fingers & Brush the left forearm with right fingertips. & 0.914 & 0.891 & 0.902 \\
48 & One heel to the other knee & Rub your right shin with your left heel. & 0.858 & 0.875 & 0.866 \\
49 & Running fingers across lips & Tap your lips with right index and middle fingers. & 0.867 & 0.906 & 0.886 \\
50 & Fist bump your hands & Press your palms together in a prayer gesture. & 0.884 & 0.869 & 0.877 \\
51 & Tracing a circle or shape on the palm with your thumb & Rub your left palm with your thumb. & 0.935 & 0.9 & 0.917 \\
52 & Smell your armpit & Smell your right armpit. & 0.97 & 0.984 & 0.977 \\
53 & Scratch your eyes with thumbs & Gently rub your eyes with both thumbs. & 0.914 & 0.93 & 0.922 \\
54 & Gently pinching bridge of nose & Pinch your nose with the right hand. & 0.917 & 0.89 & 0.903 \\
55 & Rubbing temples with fingertips & Alternating fingertip taps on temples. & 0.91 & 0.916 & 0.913 \\
56 & Running ring and middle fingertips across lips & Slide right middle and ring fingertips across your lips. & 0.919 & 0.942 & 0.93 \\
57 & Cover your mouth with your left hand over surprising news & Left hand gently touches chin and lips. & 0.882 & 0.868 & 0.875 \\
58 & Kiss your hand palm & Wipe your lips with the right palm. & 0.903 & 0.909 & 0.906 \\
59 & Send a kiss to your valentine & Right-hand smoking gesture. & 0.882 & 0.857 & 0.87 \\
60 & Crossing the arms over the chest & Cross arms over chest in a self-hug. & 0.906 & 0.939 & 0.922 \\
61 & Resting hand on chin & Hold your chin with the right hand thoughtfully. & 0.905 & 0.914 & 0.909 \\
62 & Stay in attention position & Stand with both palms on your back thighs, heels pressed together. & 0.804 & 0.83 & 0.817 \\
63 & Clean your right armpit with your left hand & Scratch your right armpit with your left hand. & 0.968 & 0.973 & 0.97 \\
64 & Tracing a shape on own forearm with the right index finger & Scratch your left forearm with your right index finger. & 0.932 & 0.911 & 0.921 \\
65 & Making a 'spider walk' motion with index and middle fingertips on the opposite forearm & Scratch your left forearm with the right index and middle fingertips. & 0.93 & 0.894 & 0.911 \\
66 & Do a squat & Do a deep squat, sitting on your heels. & 0.899 & 0.969 & 0.932 \\
67 & Holding one wrist with the opposite hand & Right hand grasps left wrist. & 0.915 & 0.91 & 0.912 \\
68 & Scratch your eyes & Rub your right eye with your right fingertips. & 0.893 & 0.924 & 0.909 \\
69 & Blow nose with your right & Pinch your nose with your right hand. & 0.935 & 0.931 & 0.933 \\
70 & Make delicious gesture & Blow a kiss with your right hand. & 0.859 & 0.85 & 0.855 \\
71 & Crossing arms & Cross arms, fingertips resting on opposite biceps. & 0.843 & 0.906 & 0.873 \\
72 & Scratching head & Gently scratch your head with right fingertips. & 0.869 & 0.863 & 0.866 \\
73 & Holding stomach & Rest both hands on your belly. & 0.871 & 0.885 & 0.878 \\
74 & Adjusting a tie & Touch chest with right fingertips. & 0.875 & 0.871 & 0.873 \\
75 & Hugging yourself & Scratching your shoulder blades with both hands. & 0.856 & 0.904 & 0.879 \\
76 & Gripping shoulder & Scratch left shoulder with right hand. & 0.87 & 0.874 & 0.872 \\
77 & Tighten your belt & Gently touch belly and waist with both hands. & 0.865 & 0.866 & 0.865 \\
78 & Salute a commander & Rub your right forehead with your right index and middle fingers. & 0.811 & 0.849 & 0.83 \\
79 & Button your trousers & Pinch your lower abdomen using right thumb and index finger. & 0.852 & 0.902 & 0.876 \\
80 & Adjusting a necklace & Straightening shirt collar with alternating hands. & 0.87 & 0.909 & 0.889 \\
81 & Rub your outer elbow & Scratch left forearm. & 0.904 & 0.88 & 0.892 \\
82 & Place hand over heart & Place right hand on chest. & 0.942 & 0.936 & 0.939 \\
83 & Gently touch your back & Scratch your mid-back with your right hand. & 0.91 & 0.916 & 0.913 \\
84 & Gently scratch your shin & Gently scratching right shin. & 0.952 & 0.946 & 0.949 \\
85 & Adjusting a tie or necklace & Lightly tap up and down your sternum with right fingertips. & 0.843 & 0.875 & 0.859 \\
86 & Running fingers along the arm & Run right fingertips down left arm. & 0.909 & 0.939 & 0.924 \\
87 & Zipping or buttoning a jacket & Pinched right hand slides upward from belly to chest. & 0.853 & 0.851 & 0.852 \\
88 & Brushing off dust from clothes & Slide your right fingertips across the chest. & 0.874 & 0.85 & 0.862 \\
89 & Roll up both sleeves one by one & Gently stroke your opposite arms. & 0.893 & 0.884 & 0.889 \\
90 & Trace a circle around your belly & Circle your navel with right index and middle fingertips. & 0.857 & 0.917 & 0.886 \\
91 & Touching chin with the left hand & Rub your chin with left fingertips. & 0.94 & 0.939 & 0.939 \\
92 & Rubbing belly with your left palm & Left-hand circular belly rub. & 0.919 & 0.906 & 0.913 \\
93 & Clap your belly with your left hand & Left palm resting on stomach. & 0.912 & 0.879 & 0.895 \\
94 & Resting one hand on the opposite shoulder & Left hand on right shoulder. & 0.938 & 0.916 & 0.927 \\
\toprule
62 & Stay in attention position & Stand with both palms on your back thighs, heels pressed together. & 0.804 & 0.83 & 0.817 \\
33 & Put your hands on your hips & Place your hands on your hips. & 0.981 & 0.977 & 0.979 \\
\toprule
\caption[\captitle]{\captitle: Pairs with lowest and the highest scores shown in the last two rows. As seen in the figure the semantic similarity between input and predicted text pairs are high, indicating the effectiveness of the high-level \gls{llm} planning.}
                        \label{tab:bert_scores_full}
                        \end{longtable}
                        \twocolumn

%% file: main.bib
@String(CVPR= {IEEE Conf. Comput. Vis. Pattern Recog.})

@String(ICCV= {Int. Conf. Comput. Vis.})

@String(ECCV= {Eur. Conf. Comput. Vis.})

@String(NIPS= {Adv. Neural Inform. Process. Syst.})

@String(ICLR = {Int. Conf. Learn. Represent.})

@String(CVPR  = {CVPR})

@String(ICCV  = {ICCV})

@String(ECCV  = {ECCV})

@String(NIPS  = {NeurIPS})

@String(ICLR  = {ICLR})

@inproceedings{Loshchilov2017DecoupledWD,
  title={Decoupled Weight Decay Regularization},
  author={Ilya Loshchilov and Frank Hutter},
  booktitle={International Conference on Learning Representations},
  year={2017},
  url={https://api.semanticscholar.org/CorpusID:53592270}
}

@inproceedings{ho2021classifierfree,
title={Classifier-Free Diffusion Guidance},
author={Jonathan Ho and Tim Salimans},
booktitle={NeurIPS 2021 Workshop on Deep Generative Models and Downstream Applications},
year={2021},
url={https://openreview.net/forum?id=qw8AKxfYbI}
}

@inproceedings{
karras2024guiding,
title={Guiding a Diffusion Model with a Bad Version of Itself},
author={Tero Karras and Miika Aittala and Tuomas Kynk{\"a}{\"a}nniemi and Jaakko Lehtinen and Timo Aila and Samuli Laine},
booktitle={The Thirty-eighth Annual Conference on Neural Information Processing Systems},
year={2024},
url={https://openreview.net/forum?id=bg6fVPVs3s}
}

@inproceedings{song2021denoising,
title={Denoising Diffusion Implicit Models},
author={Jiaming Song and Chenlin Meng and Stefano Ermon},
booktitle={International Conference on Learning Representations},
year={2021},
url={https://openreview.net/forum?id=St1giarCHLP}
}

@article{poole2022dreamfusion,
  author = {Poole, Ben and Jain, Ajay and Barron, Jonathan T. and Mildenhall, Ben},
  title = {DreamFusion: Text-to-3D using 2D Diffusion},
  journal = {arXiv},
  year = {2022},
}

@article{Nichol2021ImprovedDD,
  title={Improved Denoising Diffusion Probabilistic Models},
  author={Alex Nichol and Prafulla Dhariwal},
  journal={ArXiv},
  year={2021},
  volume={abs/2102.09672},
  url={https://api.semanticscholar.org/CorpusID:231979499}
}

@inproceedings{kingma2013auto,
  author       = {Diederik P. Kingma and
                  Max Welling},
  title        = {Auto-Encoding Variational Bayes},
  booktitle    = {ICLR},
  year         = {2014}
}

@INPROCEEDINGS{StyleGAN,
  author={Karras, Tero and Laine, Samuli and Aittala, Miika and Hellsten, Janne and Lehtinen, Jaakko and Aila, Timo},
  booktitle={2020 IEEE/CVF Conference on Computer Vision and Pattern Recognition (CVPR)}, 
  title={Analyzing and Improving the Image Quality of StyleGAN}, 
  year={2020},
  volume={},
  number={},
  pages={8107-8116},
  doi={10.1109/CVPR42600.2020.00813}}

@article{Rombach2021HighResolutionIS,
  title={High-Resolution Image Synthesis with Latent Diffusion Models},
  author={Robin Rombach and A. Blattmann and Dominik Lorenz and Patrick Esser and Bj{\"o}rn Ommer},
  journal={2022 IEEE/CVF Conference on Computer Vision and Pattern Recognition (CVPR)},
  year={2021},
  pages={10674-10685},
  url={https://api.semanticscholar.org/CorpusID:245335280}
}

@inproceedings{
  song2021scorebased,
  title={Score-Based Generative Modeling through Stochastic Differential Equations},
  author={Yang Song and Jascha Sohl-Dickstein and Diederik P Kingma and Abhishek Kumar and Stefano Ermon and Ben Poole},
  booktitle={International Conference on Learning Representations},
  year={2021},
  url={https://openreview.net/forum?id=PxTIG12RRHS}
}

@inproceedings{photorealistic_t2i,
author = {Saharia, Chitwan and Chan, William and Saxena, Saurabh and Lit, Lala and Whang, Jay and Denton, Emily and Ghasemipour, Seyed Kamyar Seyed and Ayan, Burcu Karagol and Mahdavi, S. Sara and Gontijo-Lopes, Raphael and Salimans, Tim and Ho, Jonathan and Fleet, David J and Norouzi, Mohammad},
title = {Photorealistic text-to-image diffusion models with deep language understanding},
year = {2022},
isbn = {9781713871088},
publisher = {Curran Associates Inc.},
address = {Red Hook, NY, USA},
booktitle = {Proceedings of the 36th International Conference on Neural Information Processing Systems},
articleno = {2643},
numpages = {16},
location = {New Orleans, LA, USA},
series = {NIPS '22}
}

@INPROCEEDINGS {DOODL,
author = { Wallace, Bram and Gokul, Akash and Ermon, Stefano and Naik, Nikhil },
booktitle = { 2023 IEEE/CVF International Conference on Computer Vision (ICCV) },
title = {{ End-to-End Diffusion Latent Optimization Improves Classifier Guidance }},
year = {2023},
volume = {},
ISSN = {},
pages = {7246-7256},
doi = {10.1109/ICCV51070.2023.00669},
url = {https://doi.ieeecomputersociety.org/10.1109/ICCV51070.2023.00669},
publisher = {IEEE Computer Society},
address = {Los Alamitos, CA, USA},
month =Oct}

@inproceedings{rempeluo2023tracepace,
    author={Rempe, Davis and Luo, Zhengyi and Peng, Xue Bin and Yuan, Ye and Kitani, Kris and Kreis, Karsten and Fidler, Sanja and Litany, Or},
    title={Trace and Pace: Controllable Pedestrian Animation via Guided Trajectory Diffusion},
    booktitle={Conference on Computer Vision and Pattern Recognition (CVPR)},
    year={2023}
}

@InProceedings{ghosh2022imos,
title={IMoS: Intent-Driven Full-Body Motion Synthesis for Human-Object Interactions},
author={Ghosh, Anindita and Dabral, Rishabh and Golyanik, Vladislav and Theobalt, Christian and Slusallek, Philipp},
booktitle={Eurographics},
year={2023}
}

@inproceedings{taheri2024grip,
  title  = {{GRIP}: Generating Interaction Poses Using Latent Consistency and Spatial Cues},
  author = {Omid Taheri and Yi Zhou and Dimitrios Tzionas and Yang Zhou and Duygu Ceylan and Soren Pirk and Michael J. Black},
  booktitle = {International Conference on 3D Vision ({3DV})},
  year = {2024},
  url = {https://grip.is.tue.mpg.de}
}

@inproceedings{xu2023interdiff,
   title={InterDiff: Generating 3D Human-Object Interactions with Physics-Informed Diffusion},
   author={Xu, Sirui and Li, Zhengyuan and Wang, Yu-Xiong and Gui, Liang-Yan},
   booktitle={ICCV},
   year={2023},
}

@inproceedings{zhang22couch,
title = {COUCH: Towards Controllable Human-Chair Interactions},
author = {Zhang, Xiaohan and Bhatnagar, Bharat Lal and Guzov, Vladimir and Starke, Sebastian and Pons-Moll, Gerard},
booktitle = {European Conference on Computer Vision ({ECCV})},
month = {October},
organization = {{Springer}},
year = {2022}
}

@InProceedings{Pi_2023_ICCV,
    author    = {Pi, Huaijin and Peng, Sida and Yang, Minghui and Zhou, Xiaowei and Bao, Hujun},
    title     = {Hierarchical Generation of Human-Object Interactions with Diffusion Probabilistic Models},
    booktitle = {Proceedings of the IEEE/CVF International Conference on Computer Vision (ICCV)},
    month     = {October},
    year      = {2023},
    pages     = {15061-15073}
}

@misc{ron2025hoidinihumanobjectinteractiondiffusion,
      title={HOIDiNi: Human-Object Interaction through Diffusion Noise Optimization}, 
      author={Roey Ron and Guy Tevet and Haim Sawdayee and Amit H. Bermano},
      year={2025},
      eprint={2506.15625},
      archivePrefix={arXiv},
      primaryClass={cs.CV},
      url={https://arxiv.org/abs/2506.15625}, 
}

@inproceedings{christen2024diffh2o,
      title={DiffH2O: Diffusion-based synthesis of hand-object interactions from textual descriptions},
      author={Christen, Sammy and Hampali, Shreyas and Sener, Fadime and Remelli, Edoardo and Hodan, Tomas and Sauser, Eric and Ma, Shugao and Tekin, Bugra},
      booktitle={SIGGRAPH Asia 2024 Conference Papers},
      year={2024}
}

@article{AI_Choreographer,
  title={AI Choreographer: Music Conditioned 3D Dance Generation with AIST++},
  author={Ruilong Li and Sha Yang and David A. Ross and Angjoo Kanazawa},
  journal={2021 IEEE/CVF International Conference on Computer Vision (ICCV)},
  year={2021},
  pages={13381-13392},
  url={https://api.semanticscholar.org/CorpusID:236882798}
}

@article{tseng2022edge,
  title={EDGE: Editable Dance Generation From Music},
  author={Tseng, Jonathan and Castellon, Rodrigo and Liu, C Karen},
  journal={arXiv preprint arXiv:2211.10658},
  year={2022}
}

@inproceedings{omnicontrol,
      title={OmniControl: Control Any Joint at Any Time for Human Motion Generation},
      author={Yiming Xie and Varun Jampani and Lei Zhong and Deqing Sun and Huaizu Jiang},
      booktitle={The Twelfth International Conference on Learning Representations},
      year={2024}
}

@article{DNO,
  title={Optimizing Diffusion Noise Can Serve As Universal Motion Priors},
  author={Korrawe Karunratanakul and Konpat Preechakul and Emre Aksan and Thabo Beeler and Supasorn Suwajanakorn and Siyu Tang},
  journal={2024 IEEE/CVF Conference on Computer Vision and Pattern Recognition (CVPR)},
  year={2023},
  pages={1334-1345},
  url={https://api.semanticscholar.org/CorpusID:266362434}
}

@article{GMD,
  title={Guided Motion Diffusion for Controllable Human Motion Synthesis},
  author={Korrawe Karunratanakul and Konpat Preechakul and Supasorn Suwajanakorn and Siyu Tang},
  journal={2023 IEEE/CVF International Conference on Computer Vision (ICCV)},
  year={2023},
  pages={2151-2162},
  url={https://api.semanticscholar.org/CorpusID:258833752}
}

@inproceedings{PriorMDM,
  title={Human Motion Diffusion as a Generative Prior},
  author={Shafir, Yoni and Tevet, Guy and Kapon, Roy and Bermano, Amit Haim},
  year={2024},
  booktitle={The Twelfth International Conference on Learning Representations}
}

@inproceedings{TLControl,
author = {Wan, Weilin and Dou, Zhiyang and Komura, Taku and Wang, Wenping and Jayaraman, Dinesh and Liu, Lingjie},
title = {TLControl: Trajectory and\&nbsp;Language Control for\&nbsp;Human Motion Synthesis},
year = {2024},
isbn = {978-3-031-72912-6},
publisher = {Springer-Verlag},
address = {Berlin, Heidelberg},
url = {https://doi.org/10.1007/978-3-031-72913-3_3},
doi = {10.1007/978-3-031-72913-3_3},
booktitle = {Computer Vision – ECCV 2024: 18th European Conference, Milan, Italy, September 29–October 4, 2024, Proceedings, Part XXXVII},
pages = {37–54},
numpages = {18},
location = {Milan, Italy}
}

@inproceedings{petrovich23tmr,
    title     = {{TMR}: Text-to-Motion Retrieval Using Contrastive {3D} Human Motion Synthesis},
    author    = {Petrovich, Mathis and Black, Michael J. and Varol, G{\"u}l},
    booktitle = {International Conference on Computer Vision ({ICCV})},
    year      = 2023
}

@inproceedings{motioncritic2025,
    title={Aligning Motion Generation with Human Perceptions},
    author={Wang, Haoru and Zhu, Wentao and Miao, Luyi and Xu, Yishu and Gao, Feng and Tian, Qi and Wang, Yizhou},
    booktitle={ICLR},
    year={2025}
}

@inproceedings{bert-score,
  title={{BERTScore}: Evaluating Text Generation with {BERT}},
  author={Tianyi Zhang* and Varsha Kishore* and Felix Wu* and Kilian Q. Weinberger and Yoav Artzi},
  booktitle={ICLR},
  year={2020},
  url={https://openreview.net/forum?id=SkeHuCVFDr}
}

@article{li2024unimotion,
  author    = {Li, Chuqiao and Chibane, Julian and He, Yannan and Pearl, Naama and Geiger, Andreas and Pons-Moll, Gerard},
  title     = {Unimotion: Unifying 3D Human Motion Synthesis and Understanding},
  journal   = {arXiv preprint arXiv:2409.15904},
  year      = {2024},
}

@inproceedings{zhang2024rohm,
  title={RoHM: Robust Human Motion Reconstruction via Diffusion},
  author={Zhang, Siwei and Bhatnagar, Bharat Lal and Xu, Yuanlu and Winkler, Alexander and Kadlecek, Petr and Tang, Siyu and Bogo, Federica},
  booktitle={CVPR},
  year={2024}
}

@InProceedings{HMP, 
   author = {Duran, Enes and Kocabas, Muhammed and Choutas, Vasileios and Fan, Zicong and Black, Michael J.}, 
   title = {HMP: Hand Motion Priors for Pose and Shape Estimation From Video}, 
   booktitle = {Proceedings of the IEEE/CVF Winter Conference on Applications of Computer Vision (WACV)}, 
   month = {January}, 
   year = {2024}, 
   pages = {6353-6363} 
}

@inproceedings{he2022nemf,
    author = {He, Chengan and Saito, Jun and Zachary, James and Rushmeier, Holly and Zhou, Yi},
    title = {NeMF: Neural Motion Fields for Kinematic Animation},
    booktitle = {NeurIPS},
    year = {2022}
}

@inproceedings{rempe2021humor,
    author={Rempe, Davis and Birdal, Tolga and Hertzmann, Aaron and Yang, Jimei and Sridhar, Srinath and Guibas, Leonidas J.},
    title={HuMoR: 3D Human Motion Model for Robust Pose Estimation},
    booktitle={International Conference on Computer Vision (ICCV)},
    year={2021}
}

@inproceedings{wandr,
  title = {{WANDR}: Intention-guided Human Motion Generation},
  author = {Diomataris, Markos and Athanasiou, Nikos and Taheri, Omid and Wang, Xi and Hilliges, Otmar and Black, Michael J.},
  booktitle = {Proceedings IEEE Conference on Computer Vision and Pattern Recognition (CVPR)},
  year = {2024},
}

@inproceedings{tevet2022motionclip,
  title={Motionclip: Exposing human motion generation to clip space},
  author={Tevet, Guy and Gordon, Brian and Hertz, Amir and Bermano, Amit H and Cohen-Or, Daniel},
  booktitle={Computer Vision--ECCV 2022: 17th European Conference, Tel Aviv, Israel, October 23--27, 2022, Proceedings, Part XXII},
  pages={358--374},
  year={2022},
  organization={Springer}
}

@article{SMPL:2015,
      author = {Loper, Matthew and Mahmood, Naureen and Romero, Javier and Pons-Moll, Gerard and Black, Michael J.},
      title = {{SMPL}: A Skinned Multi-Person Linear Model},
      journal = {ACM Trans. Graphics (Proc. SIGGRAPH Asia)},
      month = oct,
      number = {6},
      pages = {248:1--248:16},
      publisher = {ACM},
      volume = {34},
      year = {2015}}

@inproceedings{SMPL-X:2019,
  title = {Expressive Body Capture: {3D} Hands, Face, and Body from a Single Image},
  author = {Pavlakos, Georgios and Choutas, Vasileios and Ghorbani, Nima and Bolkart, Timo and Osman, Ahmed A. A. and Tzionas, Dimitrios and Black, Michael J.},
  booktitle = {Proceedings IEEE Conf. on Computer Vision and Pattern Recognition (CVPR)},
  pages     = {10975--10985},
  year = {2019}
}

@inproceedings{6D_representation,
title={On the Continuity of Rotation Representations in Neural Networks},
author={Zhou, Yi and Barnes, Connelly and Jingwan, Lu and Jimei, Yang and Hao, Li},
booktitle={The IEEE Conference on Computer Vision and Pattern Recognition (CVPR)},
month={June},
year={2019}
}

@article{lbs,
author = {Lewis, J.P. and Cordner, Matt and Fong, Nickson},
year = {2000},
month = {01},
pages = {165-172},
title = {Pose Space Deformation: A Unified Approach to Shape Interpolation and Skeleton-Driven Deformation},
volume = {6},
journal = {ACM SIGGRAPH},
doi = {10.1145/344779.344862}
}

@misc{openai2024o3o4mini,
  author       = {{OpenAI}},
  title        = {Introducing o3 and o4-mini},
  year         = {2024},
  month        = {May},
  url          = {https://openai.com/index/introducing-o3-and-o4-mini/},
  note         = {Accessed: 2025-05-16}
}

@conference{AMASS,
  title = {{AMASS}: Archive of Motion Capture as Surface Shapes},
  author = {Mahmood, Naureen and Ghorbani, Nima and Troje, Nikolaus F. and Pons-Moll, Gerard and Black, Michael J.},
  booktitle = {International Conference on Computer Vision},
  pages = {5442--5451},
  month = oct,
  year = {2019},
  month_numeric = {10}
}

@inproceedings{ARCTIC,
  title = {{ARCTIC}: A Dataset for Dexterous Bimanual Hand-Object Manipulation},
  author = {Fan, Zicong and Taheri, Omid and Tzionas, Dimitrios and Kocabas, Muhammed and Kaufmann, Manuel and Black, Michael J. and Hilliges, Otmar},
  booktitle = {Proceedings IEEE Conference on Computer Vision and Pattern Recognition (CVPR)},
  year = {2023}
}

@inproceedings{GRAB,
  title = {{GRAB}: A Dataset of Whole-Body Human Grasping of Objects},
  author = {Taheri, Omid and Ghorbani, Nima and Black, Michael J. and Tzionas, Dimitrios},
  booktitle = {European Conference on Computer Vision (ECCV)},
  year = {2020},
  url = {https://grab.is.tue.mpg.de}
}

@inproceedings{emage1,
  title = {{EMAGE}: Towards Unified Holistic Co-Speech Gesture Generation via Expressive Masked Audio Gesture Modeling},
  author = {Liu, Haiyang and Zhu, Zihao and Becherini, Giorgio and Peng, Yichen and Su, Mingyang and Zhou, You and Zhe, Xuefei and Iwamoto, Naoya and Zheng, Bo and Black, Michael J.},
  booktitle = {IEEE/CVF Conf.~on Computer Vision and Pattern Recognition (CVPR)},
  month = jun,
  year = {2024},
  doi = {},
  month_numeric = {6}
}

@article{omomo,
  title={Object Motion Guided Human Motion Synthesis},
  author={Li, Jiaman and Wu, Jiajun and Liu, C Karen},
  journal={ACM Trans. Graph.},
  volume={42},
  number={6},
  year={2023}
}

@InProceedings{InterHand2.6M,  
author = {Moon, Gyeongsik and Yu, Shoou-I and Wen, He and Shiratori, Takaaki and Lee, Kyoung Mu},  
title = {InterHand2.6M: A Dataset and Baseline for 3D Interacting Hand Pose Estimation from a Single RGB Image},  
booktitle = {European Conference on Computer Vision (ECCV)},  
year = {2020}  
}

@inproceedings{Reinterhand,
  title     = {A Dataset of Relighted {3D} Interacting Hands},
  author    = {Moon, Gyeongsik and Saito, Shunsuke and Xu, Weipeng and Joshi, Rohan and Buffalini, Julia and Bellan, Harley and Rosen, Nicholas and Richardson, Jesse and Mize Mallorie and Bree, Philippe and Simon, Tomas and Peng, Bo and Garg, Shubham and McPhail, Kevyn and Shiratori, Takaaki},
  booktitle = {NeurIPS Track on Datasets and Benchmarks},
  year      = {2023},
}

@inproceedings{TCDHands,
  author          = {Ludovic Hoyet and Kenneth Ryall and Rachel McDonnell and Carol O'Sullivan},
  title           = {Sleight of Hand: Perception of Finger Motion from Reduced Marker Sets},
  booktitle       = {Proceedings of the ACM SIGGRAPH Symposium on Interactive 3D Graphics and Games},
  year            = {2012},
  pages           = {79--86},
  doi             = {10.1145/2159616.2159629}
}

@inproceedings{SAMP,
  title = {Stochastic Scene-Aware Motion Prediction},
  author = {Hassan, Mohamed and Ceylan, Duygu and Villegas, Ruben and Saito, Jun and Yang, Jimei and Zhou, Yi and Black, Michael},
  booktitle = {Proceedings of the International Conference on Computer Vision 2021},
  month = oct,
  year = {2021},
  event_name = {International Conference on Computer Vision 2021},
  event_place = {virtual (originally Montreal, Canada)},
  month_numeric = {10}
}

@InProceedings{HumanML3D,
    author    = {Guo, Chuan and Zou, Shihao and Zuo, Xinxin and Wang, Sen and Ji, Wei and Li, Xingyu and Cheng, Li},
    title     = {Generating Diverse and Natural 3D Human Motions From Text},
    booktitle = {Proceedings of the IEEE/CVF Conference on Computer Vision and Pattern Recognition (CVPR)},
    month     = {June},
    year      = {2022},
    pages     = {5152-5161}
}

@article{HOT3D,
  title={{HOT3D}: Hand and Object Tracking in {3D} from Egocentric Multi-View Videos},
  author={Banerjee, Prithviraj and Shkodrani, Sindi and Moulon, Pierre and Hampali, Shreyas and Han, Shangchen and Zhang, Fan and Zhang, Linguang and Fountain, Jade and Miller, Edward and Basol, Selen and Newcombe, Richard and Wang, Robert and Engel, Jakob Julian and Hodan, Tomas},
  journal={CVPR},
  year={2025}
}

@inproceedings{MOYO,
    title = {{3D} Human Pose Estimation via Intuitive Physics},
    author = {Tripathi, Shashank and M{\"u}ller, Lea and Huang, Chun-Hao P. and Taheri Omid and Black, Michael J. and Tzionas, Dimitrios},
    booktitle = {Conference on Computer Vision and Pattern Recognition ({CVPR})},
    pages = {4713--4725},
    year = {2023},
    url = {https://ipman.is.tue.mpg.de}
}

@inproceedings{TEACH:3DV:2022,
  title={TEACH: Temporal Action Compositions for 3D Humans},
  author={Athanasiou, Nikos and Petrovich, Mathis and Black, Michael J. and Varol, G\"{u}l },
  booktitle = {International Conference on 3D Vision (3DV)},
  month = {September},
  year = {2022}
}

@inproceedings{MDM,
title={Human Motion Diffusion Model},
author={Guy Tevet and Sigal Raab and Brian Gordon and Yoni Shafir and Daniel Cohen-or and Amit Haim Bermano},
booktitle={The Eleventh International Conference on Learning Representations },
year={2023},
url={https://openreview.net/forum?id=SJ1kSyO2jwu}
}

@inproceedings{athanasiou2024motionfix,
  title = {{MotionFix}: Text-Driven 3D Human Motion Editing},
  author = {Athanasiou, Nikos and Ceske, Alp{\'a}r and Diomataris, Markos and Black, Michael J. and Varol, G{\"u}l},
  booktitle = {SIGGRAPH Asia 2024 Conference Papers},
  year = {2024}
}

@inproceedings{zhang2023generating,
            title={T2M-GPT: Generating Human Motion from Textual Descriptions with Discrete Representations},
            author={Zhang, Jianrong and Zhang, Yangsong and Cun, Xiaodong and Huang, Shaoli and Zhang, Yong and Zhao, Hongwei and Lu, Hongtao and Shen, Xi},
            booktitle={Proceedings of the IEEE/CVF Conference on Computer Vision and Pattern Recognition (CVPR)},
            year={2023},
          }

@inproceedings{ACTOR:ICCV:2021,
  title = {Action-Conditioned {3D} Human Motion Synthesis with Transformer {VAE}},
  author = {Petrovich, Mathis and Black, Michael J. and Varol, G\"{u}l},
  booktitle = {Proc. International Conference on Computer Vision (ICCV)},
  pages = {10965--10975},
  publisher = {IEEE},
  address = {Piscataway, NJ},
  month = oct,
  year = {2021},
  doi = {10.1109/ICCV48922.2021.01080},
  month_numeric = {10}
}

@article{SINC:2023,
title = {{SINC}: Spatial Composition of {3D} Human Motions for Simultaneous Action Generation},
author = {Athanasiou, Nikos and Petrovich, Mathis and Black, Michael J. and Varol, G{\"u}l },
journal = {ICCV},
year = {2023}
}

@inproceedings{taheri2021goal,
    title = {{GOAL}: {G}enerating {4D} Whole-Body Motion for Hand-Object Grasping},
    author = {Taheri, Omid and Choutas, Vasileios and Black, Michael J. and Tzionas, Dimitrios},
    booktitle = {Conference on Computer Vision and Pattern Recognition ({CVPR})},
    year = {2022},
    url = {https://goal.is.tue.mpg.de}
}
